\documentclass{article} % For LaTeX2e
\usepackage{iclr2026_conference,times}

\usepackage{amsmath,amsfonts,bm}

\def\eqref#1{equation~\ref{#1}}
\def\1{\bm{1}}

\DeclareMathAlphabet{\mathsfit}{\encodingdefault}{\sfdefault}{m}{sl}
\SetMathAlphabet{\mathsfit}{bold}{\encodingdefault}{\sfdefault}{bx}{n}

\usepackage[table]{xcolor}

\usepackage{hyperref}
\usepackage{url}
\usepackage{multirow}
\usepackage{graphicx} % For rotating text
\usepackage{lscape}   % For landscape page orientation (optional, if needed)
\usepackage{array}    % For fine-tuning column spacing
\usepackage{algorithm}
\usepackage{algpseudocode}
\usepackage{booktabs}
\usepackage{cleveref}
\usepackage{booktabs}

\usepackage{adjustbox}
\usepackage{rotating}
\usepackage{amsmath}
\usepackage{geometry}
\usepackage{threeparttable}
\usepackage{multirow}
\usepackage{makecell} % for robust multirow content

\usepackage{wrapfig, booktabs}
\usepackage{xspace}

\newcommand{\ourmethod}{\textsc{SPACeR}\xspace}

\title{SPACeR: Self-Play Anchoring with Centralized Reference Models}

\renewcommand{\thefootnote}{\fnsymbol{footnote}}

\author{%
\begin{minipage}{\textwidth}
\raggedright
Wei-Jer Chang$^{1,2}$\footnotemark[1],\;
Akshay Rangesh$^{1}$,\;
Kevin Joseph$^{1,3}$\footnotemark[1],\\
{Matthew Strong}$^{1,4}$\footnotemark[1],\;
{Masayoshi Tomizuka}$^{2}$,\;
{Yihan Hu}$^{1}$,\;
{Wei Zhan}$^{1,2}$\footnotemark[2]\\[2pt]
{\normalfont
$^1$ Applied Intuition \hspace{1em}
$^2$ University of California, Berkeley\\
$^3$ New York University \hspace{1em}
$^4$ Stanford University
}
\end{minipage}
}
\iclrfinalcopy
\begin{document}
\maketitle

\begingroup
\renewcommand{\thefootnote}{\fnsymbol{footnote}}% ensure symbol style here too
\begingroup
\renewcommand{\thefootnote}{\fnsymbol{footnote}}
\footnotetext[1]{Work done during internship at Applied Intuition. \hspace{1em}
\textsuperscript{†}Correspondence: \texttt{wei.zhan@applied.co}}
\endgroup

\begin{abstract}
Developing autonomous vehicles (AVs) requires not only safety and efficiency, but also realistic, human-like behaviors that are socially aware and predictable. Achieving this requires sim agent policies that are human-like, fast, and scalable in multi-agent settings. Recent progress in imitation learning with large diffusion-based or tokenized models has shown that behaviors can be captured directly from human driving data, producing realistic policies. However, these models are computationally expensive, slow during inference, and struggle to adapt in reactive, closed-loop scenarios. In contrast, self-play reinforcement learning (RL) scales efficiently and naturally captures multi-agent interactions, but it often relies on heuristics and reward shaping, and the resulting policies can diverge from human norms. We propose human-like self-play, a framework that leverages a pretrained tokenized autoregressive motion model as a centralized reference policy to guide decentralized self-play. The reference model provides likelihood rewards and KL divergence, anchoring policies to the human driving distribution while preserving RL scalability. Evaluated on the Waymo Sim Agents Challenge, our method achieves competitive performance with imitation-learned policies while being up to 10× faster at inference and 50× smaller in parameter size than large generative models. In addition, we demonstrate in closed-loop ego planning evaluation tasks that our sim agents can effectively measure planner quality with fast and scalable traffic simulation, establishing a new paradigm for testing autonomous driving policies.
\end{abstract}

\section{INTRODUCTION}

Developing autonomous vehicles (AVs) that can safely and smoothly share the road with human drivers is a fundamental challenge. The difficulty lies not only in ensuring safety and efficiency, but also in producing human-like behavior: policies must be predictable, socially aware, and capable of interacting effectively in complex multi-agent environments. To achieve this, it is essential to build realistic simulation policies that AVs can interact with in scalable, closed-loop testing before deployment on real roads.

%worry a bout terminology later
Simulation policies must satisfy two key requirements: realism and reactivity. Realism refers to human-like behavior, while reactivity captures meaningful responses to other agents. These properties enable AVs to handle interactive, multi-agent scenarios. Broadly, there are two paradigms for constructing such policies: imitation learning and self-play reinforcement learning (RL).

Imitation learning starts from expert demonstrations, with recent advances including tokenized models \cite{philion2024trajeglish, wu2024smart, zhang2024closed_catk} and diffusion models \cite{ctg++, jiang2024scenediffuser} to generate realistic multi-agent behaviors. The key advantage of imitation learning is that it learns directly from the human data distribution, producing realistic behavior. However, such approaches struggle in reactive settings and often require computationally heavy models such as Transformer-based architectures that limit scalability, as we demonstrate in \cref{exp:policy_eval}.

%do we want to highlgiht is centralized

%self-plaay, interactive, train agents in closed-loop interaction, however--> 1)either need a lot of heuristics 2)reward-shaping, difficulty alinging with human behaviors,
On the other hand, self-play reinforcement learning (RL) has become popular \cite{cusumano2025robust, kazemkhani2024gpudrive} for building driving policies. In self-play, agents learn policies by repeatedly playing against one another in closed-loop interaction. However, successful training often requires extensive heuristics and careful reward shaping, and the resulting policies can still diverge from human norms, leading to unrealistic and non-human like behaviors.

%credit assignment, how to merge with data,.., 
To address both challenges, we introduce \ourmethod, which leverages a pretrained tokenized model \cite{wu2024smart} to provide a reference policy distribution and a likelihood signal over the outputs of self-play policies (\cref{fig:overview}). This serves as an on-policy reward provider, shaping learning toward human-like behavior while preserving the scalability of self-play RL. Specifically, we align the self-play policy’s action space with that of the tokenized model, enabling tractable likelihood estimation and KL-based distributional alignment. This design eliminates the need for ground-truth logged trajectories and instead provides probabilistic estimates that directly guide learning.

We validate our approach on the Waymo Sim Agents Challenge (WOSAC)~\cite{montali2023wosac}, where \ourmethod significantly improves realism and human-likeness compared to prior self-play RL methods. In addition to benchmark performance, we perform closed-loop policy evaluation across diverse planners, using the reference tokenized model as a baseline. Our experiments show that Human-Like Self-Play policies are more reactive and avoid the false-positive collisions often seen in imitation-based approaches, yielding more realistic and reliable estimates for planner evaluation.

Beyond performance, the resulting policies are remarkably lightweight: decentralized MLPs with only $\sim$65k parameters that run over 10$\times$ faster and are up to 50$\times$ smaller than most tokenized models. This efficiency enables scalable, real-time multi-agent simulation at unprecedented scale, while maintaining the realism necessary for AV testing and deployment. 

\section{Related Works}

%RL

%Self-play
\textbf{Self-Play RL for autonomous driving} 

% Self-play reinforcement learning\cite{carroll2019utility_overcook, yu2022surprising_ppo, chen2024quantifying} has gained traction in autonomous driving since it allows agents to learn directly from closed-loop multi-agent interaction experience \cite{assymetric_selfplay, kazemkhani2024gpudrive, cusumano2025robust}. 
Self-play reinforcement learning has been studied extensively in multi-agent reinforcement learning (MARL) as a way for agents to learn by interacting with copies of themselves \cite{carroll2019utility_overcook, yu2022surprising_ppo, chen2024quantifying} and has recently been applied to autonomous driving \cite{assymetric_selfplay, kazemkhani2024gpudrive, cusumano2025robust}. One representative work, GIGAFlow~\cite{cusumano2025robust} demonstrates that large-scale self-play can produce robust autonomous driving policies, showcasing the scalability and effectiveness of self-play for autonomy. Nevertheless, one of the main challenges for self-play methods in real-world deployment is the divergence from human behaviors. Policies that maximize rewards may behave unpredictably, and be unable to coordinate with humans safely.
While Human-Regularized PPO~\cite{cornelisse2024human} introduces a small amount of demonstration data as a regularizer to bias policies toward human-likeness, our approach leverages a large-scale tokenized model to provide a human-like reward signal during multi-agent learning, making it the first to demonstrate competitive performance with state-of-the-art imitation learning policies while preserving the scalability of self-play.

% most previous works on self-play only focus on coll, offroad, or limited in ADe but not-humanlike -->can highlight this part too

% self-play RL, highlgiht difference to GIGAFLOW or GPUDRive , previous work, closely related to human-regularized PPO, but highlight the core gap , only measured through goal-reach

\textbf{Imitation-Learning Based Traffic simulation}
%tokenized models

Recently, data-driven methods like imitation learning have gained popularity due to their capacity to learn from an expert data-distribution with minimal human effort. Early works have explored generative imitation-learning approaches for modeling human traffic behaviors~\cite{kuefler2017imitating,bhattacharyya2020modeling}. Recent works can mainly be categorized as diffusion models and tokenized models. Diffusion Models such as ~\cite{ctg++,chang2024safeSim, langtraj_2025_ICCV, jiang2024scenediffuser} offer flexibility and controllability to potentially generate long-tail driving scenarios. On the other hand, tokenized models such as SMART~\cite{wu2024smart} or CAT-K~\cite{zhang2024closed_catk} have become more popular due to their high realism and capacity to simulate realistic multi-agent behavior. However, both approaches are computationally expensive at inference time: diffusion models require multiple denoising steps, and tokenized models generate actions sequentially. This limits scalability in large-scale closed-loop simulation. In this work, we instead leverage pretrained imitation-learning models to guide self-play RL toward human-like behaviors.
There are several works that focus on RL fine-tuning of pretrained imitation-learning models \cite{rl_finetune_waymo, chen2025riftclosedlooprlfinetuning,rlhf}: for example, using Group Relative Proximal Optimization to improve realism and controllability \cite{chen2025riftclosedlooprlfinetuning}, leveraging human feedback for post-training alignment \cite{rlhf}, or exploiting implicit preferences from pre-training demonstrations to avoid costly human annotations \cite{tian2025direct}. 
In contrast to this pretrain-then-finetune paradigm, we take an RL-first approach, where self-play serves as the foundation and imitation-learning models are incorporated as a reward provider. 

%potentially review some IRL methods

\begin{figure*}[t]
    \centering
    % Replace the path below with your actual figure file
    \includegraphics[]{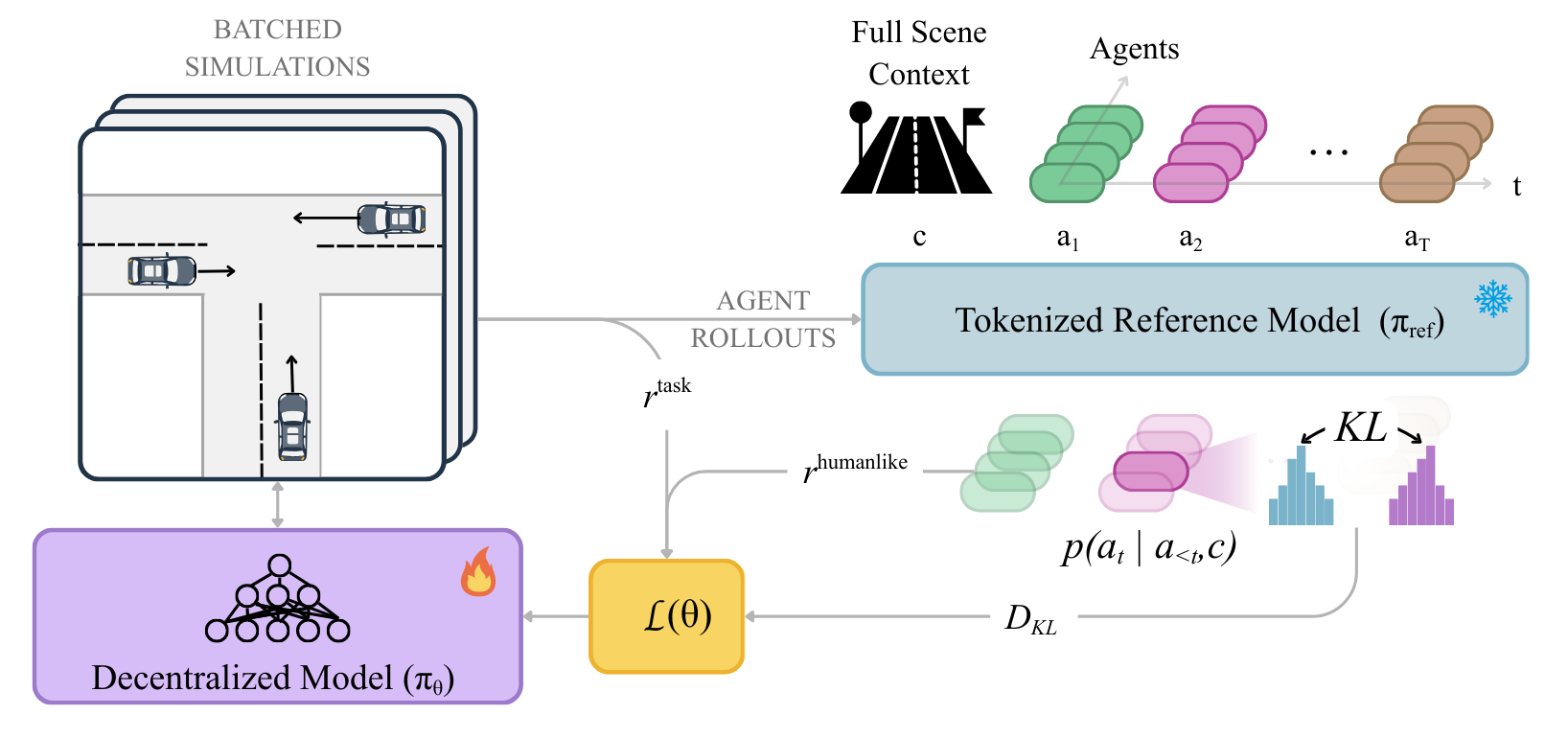}
    % If you want to add a placeholder box instead of the image, uncomment:
    % \fbox{\rule{0pt}{2in} \rule{\textwidth}{0pt}}
    \caption{\textbf{Overview of \ourmethod Framework.} 
    Self-play reinforcement learning is anchored to a pretrained tokenized 
reference model $\pi_{\text{ref}}$, which provides a human-likeness 
distributional signal. The self-play policy $\pi_{\theta}$ is decentralized 
and conditioned on local observations, whereas $\pi_{\text{ref}}$ is 
centralized and conditioned on the full scene context.}
    \label{fig:overview}
\end{figure*}

% also mention recent years, centralized models become more ,...,qcnet,..MTR

%review Pretraining as a reward model (VIPER and its following world models)?

%do we want to mention speed again here the speed?>

% \textbf{Reinforcement Finetuning of Behavior Models}

%Some work on GRPO, SMART, preference based finetuning  or do we merge it with above?

%should porbably also mention som of the RL works ( not self-play
%highlight the limitation of 1)RL 2)Traffic Simulation

%highlgiht some of the imitation+RL work from Waymo

%IRL
\section{Human-like self-play}\label{sec:method}
We aim to learn a human-like driving policy $\pi_{\theta}(a|o)$ whose behaviors match the real human driving distribution in closed-loop. The desired policy should reach goals, avoid collisions, and stay on-road, while producing predictable and realistic behaviors.

\subsection{Problem Formulation}
We formulate this as a partially observable multi-sequential decision-making problem. 
At each timestep $t$, the global simulator state $s_t \in \mathcal{S}$ encodes 
the static road graph and the dynamic states of all agents. Each agent $i$ 
receives only a partial observation $o_t^i = \mathcal{O}(s_t, i)$ within its 
local field-of-view. The agent then chooses an action $a_t^i \in \mathcal{A}$ (e.g., acceleration, steering, or 
maneuver) according to its policy $\pi_{\theta}(a_t^i|o_t^i)$, and the environment 
transitions according to $s_{t+1} \sim T(s_t, a_t^1,\dots,a_t^N)$.

The key challenge in training a human-like policy $\pi_{\theta}$ is that task-based 
rewards (e.g., reaching goals, avoiding collisions, or staying on the road) 
encourage efficiency and safety but do not guarantee realism. Such rewards 
often lead to behaviors that maximize task success while diverging from 
human driving norms, for example, accelerating unnaturally to reach goals 
or performing overly sharp maneuvers to avoid collisions. To address this, 
we introduce a pretrained reference policy $\pi_{\text{ref}}$ that captures 
the human driving distribution and provides a realism signal during 
self-play, anchoring learning to behaviors observed in real data. 
%mention the PPO objective first
% reward shaping
\begin{equation}
r_t \;=\; r^{\text{task}}_t \;+\; \alpha \, r^{\text{humanlike}}(s_t,a_t),
\label{eq:reward}
\end{equation}

% abstract PPO objective with KL regularizer
\begin{equation}
\mathcal{L}(\theta)
= \mathcal{L}_{\mathrm{PPO}}(\theta; A[r])
- \beta \, D_{\mathrm{KL}}\!\big(\pi_{\theta}(\cdot \mid o_t)\,\|\,\pi_{\mathrm{ref}}(\cdot \mid s_t)\big),
\label{eq:ppo}
\end{equation}

Eq.~\ref{eq:ppo} combines three components:

\begin{itemize}
    \item \textbf{Task performance:} 
    $\mathcal{L}_{\mathrm{PPO}}(\theta)$ is the standard PPO objective \cite{schulman2017ppo}, 
    encouraging agents to reach goals, avoid collisions, and stay on the road.

    \item \textbf{Human-likeness reward:}  
    Inspired by prior work~\cite{viper}, we define a log-likelihood based reward that provides dense, per-timestep feedback by scoring each executed action under the reference distribution:  
    \begin{equation}
    r_{\mathrm{humanlike}}(s_t, a_t) = \log \pi_{\mathrm{ref}}(a_t \mid s_t).
    \end{equation}
    This formulation corresponds to maximizing the likelihood of human-like actions, thereby encouraging realistic driving behaviors beyond task success.

    \item \textbf{Distributional alignment:} 
    The KL divergence 
    $D_{\mathrm{KL}}(\pi_{\theta}(\cdot \mid o_t)\,\|\,\pi_{\mathrm{ref}}(\cdot \mid s_t))$ 
    acts as a dense signal at every timestep, directly encouraging the policy 
    to align its action distribution with the human driving distribution 
    captured by the reference model. For notational simplicity, we omit the agent index $i$; the KL divergence is computed independently for each agent.

\end{itemize}

\subsection{Pretrained Reference Tokenized Model}
To incorporate human-likeness into self-play, we introduce a pretrained reference policy $\pi_{\text{ref}}$, trained on real-world human driving trajectories, to serve as a proxy for the human driving distribution. 
Unlike task-based rewards, which only encourage efficiency and safety, 
$\pi_{\text{ref}}$ provides a distributional signal that anchors policies 
toward realistic behavior. In principle, $\pi_{\text{ref}}$ can be any 
data-driven generative motion model, such as a diffusion model or a 
tokenized model. In this work, we focus on tokenized models 
(e.g., SMART~\cite{wu2024smart}, CAT-K~\cite{zhang2024closed_catk}), since they provide tractable likelihoods estimates suitable for distributional alignment.
%direct KL

\textbf{Centralized Reference Tokenized Model.}
Tokenized models such as SMART~\cite{wu2024smart} operate in an 
autoregressive fashion: given the past joint actions $a_{<t}$ and the 
scene context $c$ (e.g., road graph and initial states), the model predicts 
a distribution over the next joint action $a_t = (a_t^1,\dots,a_t^N)$, 
where $t$ denotes the discrete timestep and $N$ is the number of agents 
in the scene. Under a conditional independence assumption across agents, 
this factorizes as
\begin{equation}
p(a_t \mid a_{<t}, c) 
= \prod_{i=1}^N p(a_t^i \mid a_{<t}, c).
\end{equation}
In our framework, we treat $p(\cdot)$ as the pretrained reference policy 
$\pi_{\mathrm{ref}}$, which provides a distinct distribution for each 
agent’s action $a_t^i$ at every timestep, conditioned on the shared scene 
context and action history.

% mention that we asign different distribution
Importantly, the ability of $\pi_{\mathrm{ref}}$ to assign a distinct 
distribution to each agent’s action at every timestep directly addresses 
a central challenge in multi-agent reinforcement learning: the credit 
assignment problem. It is often unclear which agent, and at which 
timestep, is responsible for a positive or negative outcome. This fine-grained feedback, grounded in the distributional signal of 
$\pi_{\mathrm{ref}}$, enables shaping learning signals on a per-agent, 
per-timestep basis rather than relying solely on sparse, trajectory-level 
rewards.

% One key challenges in multi-agent reinforcement learning is the 
% credit assignment problem: it is often unclear which agent, and at which 
% timestep, is responsible for a positive or negative outcome. In our 
% formulation, the reference tokenized model $\pi_{\mathrm{ref}}$ naturally 
% alleviates this issue by providing likelihood estimates for each agent’s 
% action at every timestep. This fine-grained feedback enables us to shape 
% learning signals on a per-agent, per-timestep basis, rather than relying 
% solely on sparse, trajectory-level rewards.

\textbf{(1) Tractable training without autoregressive sampling.} 
Unlike autoregressive generation, which requires sequential token 
sampling, our approach only requires a single forward pass of the 
reference model per rollout batch. This provides the full 
per-agent, per-timestep action distribution, making the training 
pipeline efficient and scalable. 

\textbf{(2) Aligned action space.} 
By ensuring that $\pi_{\theta}$ and $\pi_{\mathrm{ref}}$ operate over the same 
discrete action space, we avoid online tokenization during training. 
This alignment not only reduces computational overhead but also 
enables the KL divergence to be computed directly in closed form:
% \[
% D_{\mathrm{KL}}\!\big(\pi_{\theta}(\cdot \mid o_t)\,\|\,\pi_{\mathrm{ref}}(\cdot \mid s_t)\big) 
% = \sum_{a \in \mathcal{A}} \pi_{\theta}(a \mid s_t)\,\log 
% \frac{\pi_{\theta}(a \mid o_t)}{\pi_{\mathrm{ref}}(a \mid s_t)} .
% \]

\begin{equation}
D_{\mathrm{KL}}\!\big(\pi_{\theta}(\cdot \mid o_t)\,\|\,\pi_{\mathrm{ref}}(\cdot \mid s_t)\big) 
=
\sum_{a \in \mathcal{A}} 
\pi_{\theta}(a \mid o_t)\,
\log 
\frac{\pi_{\theta}(a \mid o_t)}
{\pi_{\mathrm{ref}}(a \mid s_t)} .
\label{eq:kl}
\end{equation}

\textbf{(3) Privileged information and comparison to log-replay data.}
 Note that in Eq.~\ref{eq:ppo}, the reference tokenized model is centralized, 
observing the full state of all agents rather than local views. 
This setup resembles privileged information in a teacher–student framework in Robotics and Computer Vision~\cite{he2024omnih2o, caron2021emerging}.  
Unlike log replay, the reference model generalizes beyond recorded trajectories and provides guidance in unseen self-play states.

\textbf{Reference Models vs. WOSAC Metrics.}
WOSAC metrics~\cite{wu2024smart} estimate per-feature likelihoods
(kinematic, interactive, and map-based) based on ground-truth future trajectories. 
In contrast, tokenized reference models directly define a distribution 
over entire action sequences, offering two advantages: (i) evaluation does not 
require logged future trajectories, and (ii) the model provides a principled 
sequence-level distribution rather than feature-wise likelihoods tied to manual design metrics.
%toekn space, frequency, maybe should highlgiht

% \subsection{Common problems of RL settings, details,..,}
\subsection{Practical Considerations of building human-like self-play}
Here, we outline common problems in standard RL settings and the 
corresponding considerations for building human-like self-play.

%the problem of goal-conditioning in RL, and behaviors after reaching goal
\textbf{Goal-reaching reward and post-goal behavior}
In previous works \cite{kazemkhani2024gpudrive, reliable_simagent}, 
agents are typically rewarded only upon reaching their goal, and are 
removed from the scenario once the goal is reached. This formulation has 
two main drawbacks: (i) agents are incentivized to accelerate unnaturally 
in order to reach the goal as quickly as possible, and (ii) the number of 
active agents decreases after goals are reached, making the scenario 
progressively easier. Qualitatively, with this original formulation, if the agents are not disappeared, agents would mostly stop once reached the goal, making the scenarios unrealistic.
To address this issue, we adopt \textit{goal-dropout}: agents are trained with and without goal conditioning while receiving a terminal goal reward. This reduces reliance on explicit goal inputs. We further show that, when anchored to a reference model, the explicit goal reward can be removed without performance loss.

\textbf{Tokenized Action Space and policy frequency}:
Contrary to previous works that utilize unicycle dynamics, we align the 
action space with the reference  model by adopting a tokenized 
trajectory action space with K-disk from the data \cite{philion2024trajeglish, wu2024smart}. This ensures compatibility for computing 
likelihoods and KL divergence without online tokenization. Therefore, the simulation frequency and policy frequency may differ; for example, the policy can operate at 5 Hz while the simulator runs at 10 Hz.

% \subsection{Closed-loop Policy Evaluation}
% -How we setup policy evaluation,...
% Different speccifically to CAT-K ( Collision estimate)
% Correlation results
% -Rule-based planner (IDM, frenet planners), Self-play checkpoints
% -- toal 42 planners

% \textbf{include a discussion on VRU self-play}: if we want to include the results

%should mention the reward, and data setup

%where do we want to mention, 

% Goal conditioning leaks temporal information by giving agents their final 
% destinations, whereas in practice driving behavior should unfold based on 
% local interactions rather than predefined goal knowledge.

%goal dropouts

%tokenized action space, and sample frequency 

%dynamic batch size

% Common problems of RL settings, details,..,
%what are the challenges

% Highlgiht, the problem we are focusing on 

% 1) Problem Formulation
% TO highlgiht details:
% 1) common problems of RL: 1) reach goal dissapeared 2)goal-point reward encoruage speedup
% 2)classic dynamics verus traj token space
% 3) policy execute at 5 Hz
% 4) dynamic batch size (transitions)
% 5)How we speedup smart like reward (batched instead of sampling), the speed, how we speed it up
% small details: we also delay the goal-reaching reward to the end of the episode, how we align the action space
% implementation details of smart,
% in supp
%dynamics verus toke space
%what are the speed

% \section{Experimental Design}

\section{Experimental Results}
We validate whether self-play policies trained with 
reference model produce behaviors that are both \textit{human-like} 
and \textit{reactive}. Experiments are conducted on large-scale traffic 
scenarios from the Waymo Open Motion Dataset (WOMD) \cite{waymo_dataset}, 
measuring how closely the learned policies match the human driving 
distribution and how effectively they adapt to interactions in closed-loop 
simulation.

% \begin{figure}
%     \centering
%     \includegraphics[width=1\linewidth]{fig/comparison_fig2.png}
%     \caption{Enter Caption}
%     \label{fig:placeholder}
% \end{figure}

\subsection{Implementation Details}
% 3.1 Dataset and Simulator
We conduct all experiments in GPUDrive~\cite{kazemkhani2024gpudrive}, 
a GPU-accelerated, data-driven simulator built on WOMD~\cite{waymo_dataset}. 
Each WOMD scenario spans 9 seconds; following the setup of \cite{montali2023wosac}, 
we initialize at 1 second and simulate the remaining 8 seconds. 
We train on 10k scenarios. In the main vehicle experiments (\cref{tab:wosac}), we control only vehicles, while pedestrians and cyclists follow their logged trajectories. For the VRU experiments (\cref{tab:wosac_vru}), we train SPACeR by controlling all agent types, , but report metrics only on pedestrian/cyclist targets.

\textbf{Observation Space}. We model the RL problem as a Partially Observed Stochastic Game~\cite{posg_game}, where agents act simultaneously under partial observability. 
Each controlled vehicle receives a local observation $o_t^i$ in an ego-centric coordinate frame, 
including nearby vehicles, lane geometry, goal points (optionally), and relevant road features within a 
50\,m radius. Agents do not receive temporal history, and all features are normalized to the range $[-1, 1]$.

\textbf{Action Space}. We adopt a tokenized trajectory action space following 
\cite{philion2024trajeglish,wu2024smart}. Using the K-disk 
algorithm, we cluster short-horizon trajectories in Cartesian space into $K=200$ discrete tokens. Each token represents a 0.1-second step with a horizon length of 2, corresponding to a 5 Hz action frequency. This setup balances two considerations: providing sufficiently fine-grained distributional signals from the SMART reference model while keeping memory usage manageable. Since actions are defined directly in Cartesian space, no explicit dynamics model is assumed; the simulator advances the scenario according to the selected token.

\textbf{Reward Formulation.} For each agent, the task reward is
\[
r^{\text{task}}(s_t,a_t) \;=\;
w_{\text{goal}}\,\mathbb{I}[\text{Goal achieved}]
- w_{\text{collided}}\,\mathbb{I}[\text{Collided}]
- w_{\text{offroad}}\,\mathbb{I}[\text{Offroad}]
+ w_{\text{humanlike}}\,r^{\text{humanlike}}(s_t,a_t),
\]
where $\mathbb{I}[\cdot]\in\{0,1\}$. By default, we set $w_{\text{collided}}=w_{\text{offroad}}=0.75$. The weights $w_{\text{goal}}$ and $w_{\text{humanlike}}$ are varied in the following ablation experiments to study the trade-off between task completion, safety, and human-likeness. Note that other than the reward, we also directly compare the KL divergence between trained policy and referenced policy, where we use CAT-K \cite{zhang2024closed_catk} as the reference model. 

\textbf{Model Architecture and Training Details}: We use a late-fusion feedforward model \cite{kazemkhani2024gpudrive,reliable_simagent}, where ego, partner, and road-graph features are each embedded with a two-layer MLP and then concatenated. The fused representation is fed into an actor head and a critic head. During training, we control up to 64 agents with a shared decentralized policy $\pi_{\theta}$, where each agent samples actions based on its local observation. We optimize with Proximal Policy Optimization (PPO) \cite{schulman2017ppo}, with hyperparameters provided in \cref{sup:training_details}. Training is performed on a single NVIDIA A100 GPU for 1 billion environment steps. During training, all VRUs (pedestrians and cyclists) follow the logged trajectories except for the VRU experiments in Table~\ref{tab:wosac_vru}, where we train SPACeR for all agent types, and calculate metrics on VRUs.

% We formulate driving as a partially observable multi-agent decision-making problem, where each agent receives a local observation derived from the global simulator state

% 3.2 Observations and Actions

% 3.3 Reward Design
% Preamble (once in your paper):
% \usepackage{booktabs}
% \usepackage[table]{xcolor}
\definecolor{ILRow}{gray}{0.9} % imitation learning shading (light grey)
\newcommand{\meanstd}[2]{#1{\scriptsize\,$\pm$\,#2}}

\begin{table}[t!]
\centering
\setlength{\tabcolsep}{4pt}
\renewcommand{\arraystretch}{0.90}
\resizebox{\linewidth}{!}{%
\begin{tabular}{lcccccccc}
\toprule
\textbf{Method} & \textbf{Composite $\uparrow$} & \textbf{Kinematic $\uparrow$} 
& \textbf{Interactive $\uparrow$} & \textbf{Map $\uparrow$}
& \textbf{minADE $\downarrow$} & \textbf{Collision $\downarrow$} & \textbf{Off-road $\downarrow$} & \textbf{Throughput $\uparrow$} \\
\midrule
PPO
  & \meanstd{0.710}{0.01} & \meanstd{0.327}{0.01} & \meanstd{0.751}{0.01} & \meanstd{0.875}{0.00} & \meanstd{12.725}{2.53} & \meanstd{0.038}{0.005} & \meanstd{0.053}{0.00} & \meanstd{211.8}{5.6} \\
HF-PPO
  & \meanstd{0.716}{0.00} & \meanstd{0.341}{0.00} & \meanstd{0.756}{0.01} & \meanstd{0.880}{0.00} & \meanstd{12.254}{1.02} & \meanstd{0.044}{0.006} & \meanstd{0.053}{0.00} & \meanstd{211.8}{5.6} \\
\textbf{SPACeR}
  & \meanstd{0.741}{0.00} & \meanstd{0.411}{0.00} & \meanstd{0.779}{0.01} & \meanstd{0.880}{0.00} & \meanstd{4.101}{0.09} & \meanstd{0.036}{0.010} & \meanstd{0.056}{0.00} & \meanstd{211.8}{5.6} \\
\midrule
\rowcolor{ILRow}
SMART
  & 0.720 & 0.450 & 0.725 & 0.870 & 1.840 & 0.17 & 0.13 & \meanstd{22.5}{0.0} \\
\rowcolor{ILRow}
CAT\mbox{-}K
  & 0.766 & 0.490 & 0.792 & 0.890 & 1.470 & 0.06 & 0.09 & \meanstd{22.5}{0.0} \\
\bottomrule
\end{tabular}}
\caption{\textbf{Results on the WOSAC Validation Set.} 
Our proposed method outperforms other self-play approaches across all realism metrics, while achieving $\sim$10$\times$ higher throughput than imitation-learning (shaded) methods, with competitive performance and lower collision/off-road rates. Throughput is measured in scenarios/sec at 5 Hz on a single A100 GPU.}

\label{tab:wosac}
% \vspace{-0.4cm}
\end{table}

% 3.4 Practical Considerations
\subsection{Humanlike self-play Evaluation on WOSAC}\label{sec:exp:humanlike}
To measure humanlike behaviors, we adopt the 
evaluation protocol from the Waymo Open Sim Agent Challenge (WOSAC) 
\cite{montali2023wosac}. WOSAC
evaluates the distributional realism of simulated agents by comparing 
their rollouts against held-out human driving data, details provied in \cref{sup:metrics}. Metrics assess  kinematics (e.g., speed, acceleration), inter-agent 
interactions, and adherence to map constraints, with results aggregated 
into a composite realism score. %\cref*{sup:metrics}

\textbf{PPO (Self-Play).} 
Decentralized PPO trained only with the task reward $r_{\text{task}}$~\cite{kazemkhani2024gpudrive}; no realism signals are used.  

\textbf{HR-PPO}~\cite{cornelisse2024human}. 
PPO regularized toward a behavioral cloning (BC) reference using KL divergence only (no likelihood term). 
For fairness, the BC model shares the same backbone as $\pi_\theta$, is decentralized, and conditions only on local observations; it is trained on the WOMD training split. 

\textbf{Imitation learning baselines.} 
We evaluate two tokenized closed-loop models—\textsc{SMART}~\cite{wu2024smart} and \textsc{CAT\mbox{-}K}~\cite{zhang2024closed_catk}. 
Both use the same backbone and action vocabulary ($K\!=\!200$) and run at 5\,Hz. 
\textsc{SMART} is first trained by behavior cloning on WOMD and then fine-tuned following the \textsc{CAT\mbox{-}K} closed-loop fine-tuning protocol; 
Implementation details and hyperparameters are provided in \cref{sup:training_details}.

\begin{wrapfigure}{r}{0.50\textwidth}
    \vspace{-10pt}
    \centering
    \includegraphics[width=0.48\textwidth]{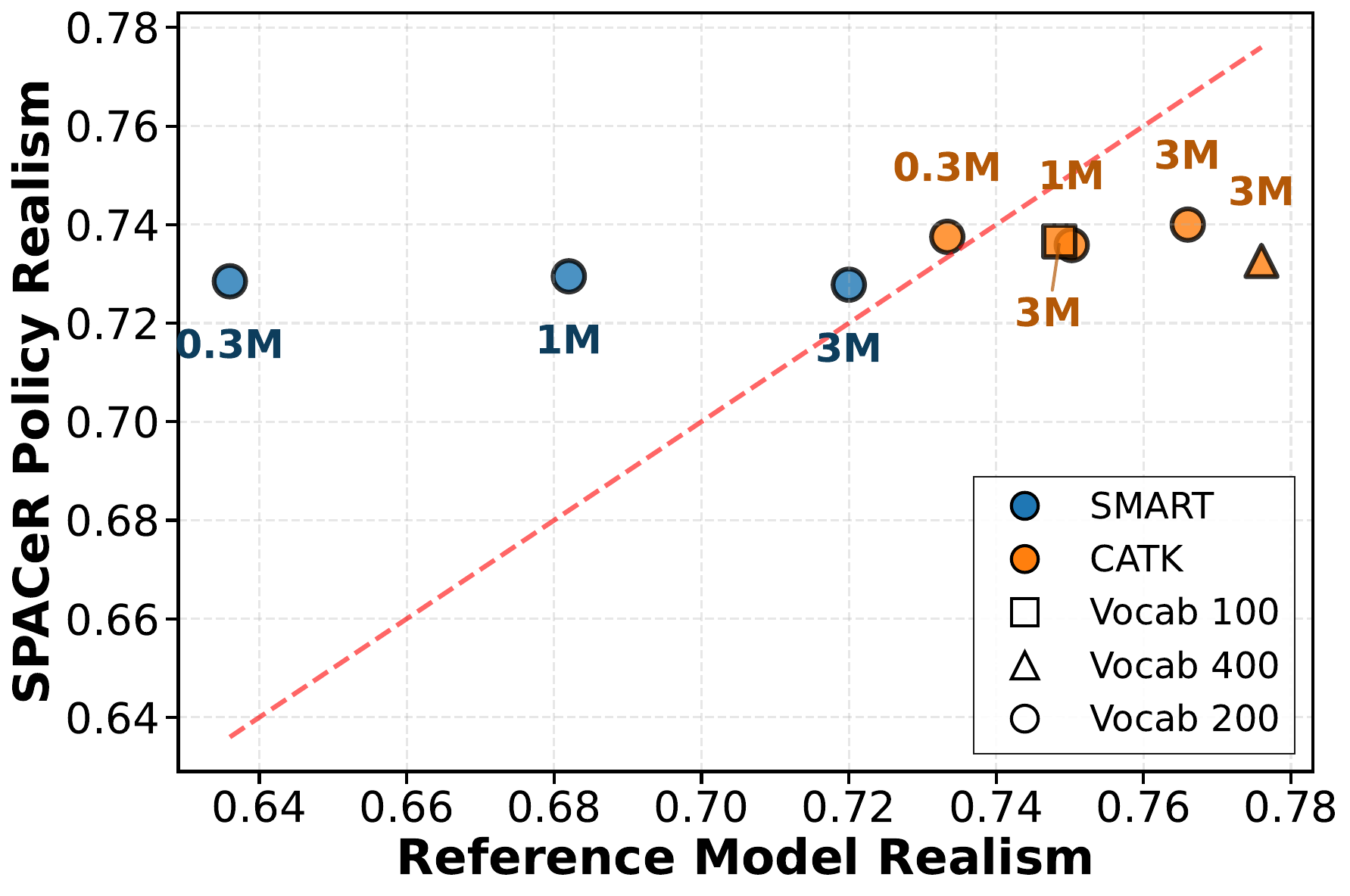}
    \vspace{-8pt}
    \caption{\textbf{Reference Model Quality Effect on SPACeR Realism.}  
    SPACeR maintains high composite realism (~0.73) even when anchored to a reference model of only 0.3 M parameters with realism 0.64.}
    \label{fig:ref_quality_scatter}
    \vspace{-8pt}
\end{wrapfigure}

\noindent\textbf{Discussion of Main Results.} 
In \cref{tab:wosac}, anchoring self-play with a pretrained reference model substantially improves realism across all Sim Agent metrics.
By contrast, HR-PPO reduces \emph{minADE} but yields only a modest realism gain, whereas our method achieves clear improvements in kinematics, underscoring the value of alignment with a strong reference model. Relative to imitation-learning methods, our approach attains lower collision and off-road rates and even surpasses \textsc{SMART} in composite realism. 
% In \cref{tab:wosac}, anchoring self-play with a pretrained reference model substantially improves realism across all Sim Agent metrics. By contrast, HR-PPO reduces \emph{minADE} but yields only a modest realism gain. 

For efficiency and scale, our decentralized MLP has $\sim$65k parameters versus $\sim$3.2M for \textsc{CAT\mbox{-}K} ($\sim$50$\times$ smaller). Briefly, we observe $\sim$10$\times$ higher closed-loop throughput (single A100, 5\,Hz); see the rightmost column of \cref{tab:wosac} and \cref{exp:policy_eval} for details. 
\textsc{CAT\mbox{-}K} remains the strongest IL baseline, but our method preserves human-likeness at substantially higher speed and greater reactivity, which we further demonstrate in the closed-loop policy evaluation (\cref{exp:policy_eval}).

% End the previous paragraph BEFORE starting wrapfigure
% -----------------------------------------------------}

\noindent\textbf{Reference Model Quality.} We further study how reference model quality affects SPACeR performance by training \textsc{SMART} models of three sizes: 0.3M, 1M, and 3M parameters. Each model is also fine-tuned with CAT-K, producing a total of six reference models (\cref{fig:ref_quality_scatter}).
SPACeR policy still reaches 0.732 realism even with the smallest underperforming reference model (0.636).
We observe a similarly stable performance when varying the token vocabulary size with 3M params (100, 200, 400; different markers in \cref{fig:ref_quality_scatter}): larger vocabularies increase reference model's realism, but the resulting SPACeR policies all remain clustered around 0.73.
This indicates that the reference model serves mainly as a soft prior that guides humanlike behavior, rather than a target for direct imitation.

Because SPACeR agents learn through closed-loop interaction, they can refine their behavior beyond the limitations of the reference model.
In addition, using a reference model that has itself undergone closed-loop fine-tuning (CAT-K) provides stronger behavioral signals, which further improves the resulting SPACeR agents performance.

\begin{table}[t]
\vspace{-4pt}  % tighten space above
\centering
\footnotesize              % larger than \scriptsize
\setlength{\tabcolsep}{4pt}% mild column shrink
\renewcommand{\arraystretch}{0.9} % only slightly tighter rows

% \begin{tabular}{lccccc}
% \toprule
% \textbf{Method} & \textbf{Composite $\uparrow$} & \textbf{Kinematic $\uparrow$} &
% \textbf{Interactive $\uparrow$} & \textbf{Map $\uparrow$} & \textbf{minADE $\downarrow$} \\
% \midrule
% PPO           & 0.628 & 0.194 & 0.670 & 0.822 & 8.755 \\
% HR-PPO     & 0.651 & 0.273 & 0.731 & 0.764 & 4.103 \\
% SPACeR (Ours) & \textbf{0.730} & \textbf{0.412} & \textbf{0.761} &
%                  \textbf{0.872} & \textbf{2.054} \\
% \bottomrule
% \end{tabular}

\begin{tabular}{lccccc}
\toprule
\textbf{Method} & \textbf{Composite $\uparrow$} & \textbf{Kinematic $\uparrow$} &
\textbf{Interactive $\uparrow$} & \textbf{Map $\uparrow$} & \textbf{minADE $\downarrow$} \\
\midrule
PPO           & 0.648 & 0.242 & 0.683 & 0.835 & 7.712 \\
HR-PPO        & 0.668 & 0.285 & 0.700 & 0.847 & 7.014 \\
SPACeR (Ours) & \textbf{0.729} & \textbf{0.413} & \textbf{0.762} &
                 \textbf{0.866} & \textbf{2.066} \\
\bottomrule
\end{tabular}
\caption{{\textbf{VRU realism metrics on WOSAC (pedestrians and cyclists).}
SPACeR outperforms PPO and HR-PPO by a large margin, achieving substantial gains across all realism metrics and minADE.}}
\label{tab:wosac_vru}
\vspace{-8pt}  % tighten space below
\end{table}

\noindent\textbf{Evaluation on Pedestrian and Cyclist Behavior.}
We evaluate SPACeR on the WOSAC validation set restricted to pedestrians and cyclists to isolate VRU behavior (\cref{tab:wosac_vru}). Although absolute composite realism scores are 0.1 lower than for vehicles, SPACeR still improves performance by a large margin over PPO and HR-PPO across all metrics, including Composite, Kinematic, Interactive, and Map realism, as well as minADE. This demonstrates that the anchoring mechanism remains effective even under the higher stochasticity and variability of VRU motion. Additional design choices and component ablations for VRU simulation are provided in Appendix \cref{supp:vru_details}.

% \textcolor{blue}{Nevertheless, VRU self-play remains underexplored in the literature. Existing WOSAC metrics such as off-road rate and collision rate are primarily designed for vehicles, while VRU behavior often requires metrics that capture sidewalk adherence and walkway usage. Reward design must likewise reflect these VRU-specific considerations. In this work we primarily use likelihood and collision signals for VRU motion. Developing VRU-aware metrics, reward shaping, and scene-level infrastructure is an important direction for enabling more realistic pedestrian and cyclist simulation.
% }

{\noindent\textbf{Effect of Anchoring Parameters on SPACeR Performance.}} 
We investigate the effect of the anchoring parameters, namely the likelihood weight $\alpha$ and the KL alignment weight $\beta$, by activating each term independently.
Top-1 refers to evaluating agents by always taking the single most probable action, whereas top-5 evaluates realism under stochastic sampling from the top five most probable actions. Since SimAgent measures distributional realism over 32 joint rollouts, maintaining diversity through top-$k$ sampling generally yields higher composite scores~\cite{montali2023wosac}.

\begin{wrapfigure}{r}{0.50\textwidth}
\vspace{-10pt}
\centering
\includegraphics[width=0.48\textwidth]{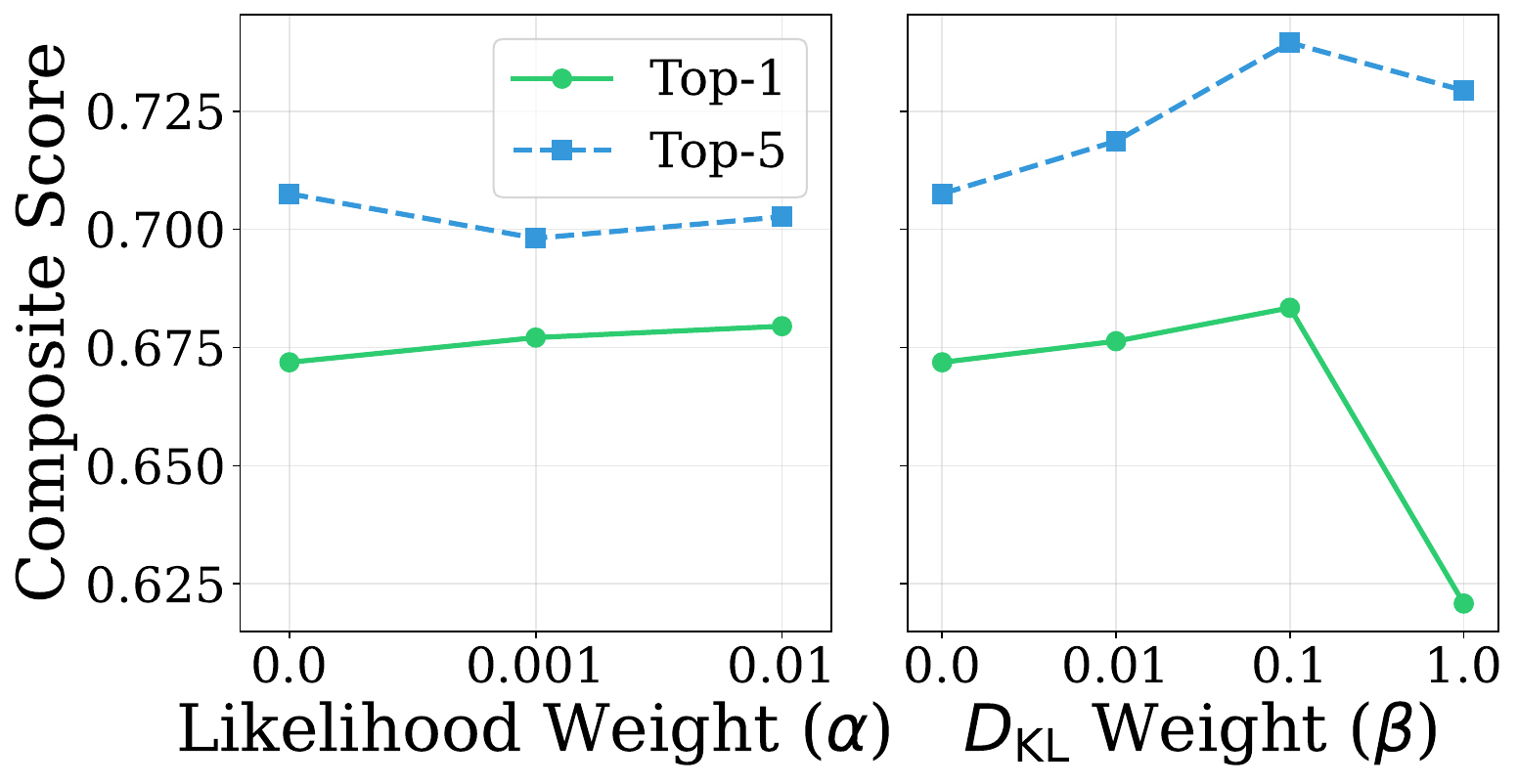}
\vspace{-6pt}
\caption{
\textbf{Anchoring parameter ablation.}
Likelihood-only improves top-1, while KL alignment improves both top-1 and top-5 realism.
}
\label{fig:anchor_ablation_small}
\vspace{-5pt}
\end{wrapfigure}

The likelihood and KL alignment terms play different roles in shaping policy behavior. Optimizing the likelihood term alone increases log likelihood but reduces diversity, as reflected by a drop in entropy in the training curves (Appendix~\cref{fig:weight_ablation}). This produces small improvements in top-1 composite realism but harms top-5 performance. In contrast, the KL alignment term raises both likelihood and KL consistency while preserving entropy, which improves both top-1 and top-5 realism. The training curves in Appendix~\cref{fig:weight_ablation} illustrate how log likelihood, KL divergence, and entropy evolve under each parameter setting.

In the ablation study (\cref{tab:wosac_ablate}), KL alignment contributes most to realism. While log-likelihood rewards are popular~\cite{viper}, their impact on realism is modest, likely because the real-world driving distribution is highly multi-modal, so maximizing likelihood alone does not yield a stable or sufficiently discriminative signal. 
A second insight is that once the policy is anchored to the reference distribution, the explicit goal-reaching reward can be removed, further improving realism.

\subsection{\ourmethod Agents for Fast \& Accurate Closed-Loop Planner Evaluation}\label{exp:policy_eval}

\begin{figure}[t]
    \centering
    \includegraphics[width=0.9\textwidth, keepaspectratio]{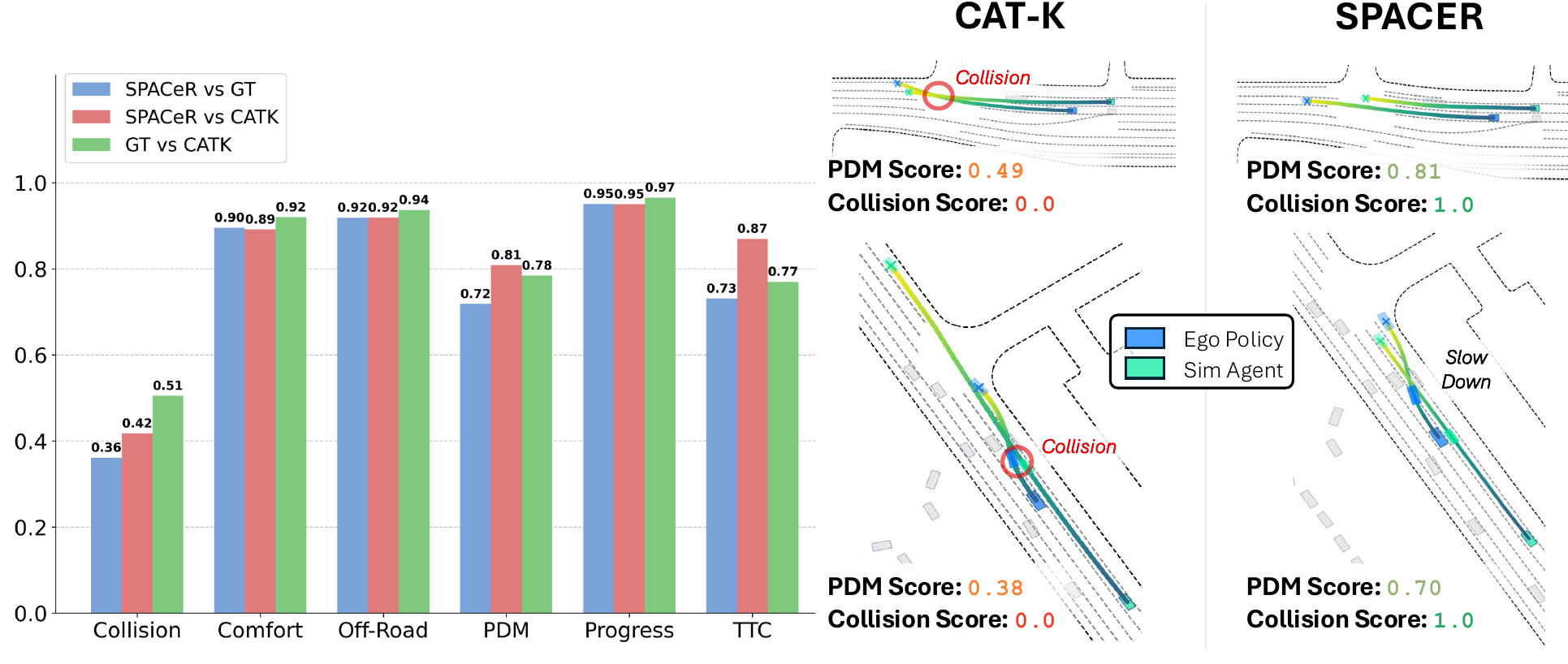}
    \caption{\textbf{Correlation Coefficients of PDM Scores Across Policy Evaluation Strategies.} Using our approach leads to consistently lower correlations with ground-truth log replays across all metrics, suggesting more realistic penalization of unsafe policies—especially for collisions.}
    \label{fig:policy_evaluation}
\end{figure}

A key application of humanlike agents is the closed-loop evaluation of ego-vehicle motion planners. In this setup, the planner under test controls the ego vehicle, while surrounding agents are driven by learned policies. This enables realistic modeling of inter-agent interactions, reactive behaviors, and humanlike decision-making without risking safety.

We evaluate our \ourmethod agents along two dimensions: 
\textit{closed-loop evaluation realism} and \textit{throughput}. 
Unlike distributional realism in \cref{sec:exp:humanlike}, which measures similarity to human data, 
realism here refers to how sim agents change planner rankings under closed-loop interaction.

To assess realism, we compare against two baselines: (i) open-loop log-replay evaluation, and (ii) closed-loop simulation using CAT-K \cite{zhang2024closed_catk}, a state-of-the-art traffic simulator on WOSAC. We evaluate a diverse set of planners: 18 self-play–trained policies, 10 sampling-based Frenet planners, and 10 IDM-based planners~\cite{treiber2000congested} (more details can be found in~\ref{sup:planner_variant}). For each planner, we compute PDM scores~\cite{dauner2024navsim} across multiple scenes under three traffic-simulation strategies: ground-truth logs (GT), CAT-K rollouts, and \ourmethod agent policies. We then measure the correlation of PDM scores between strategies. As shown in \cref{fig:policy_evaluation}, our approach yields consistently lower correlations with GT replays when compared to CAT-K. We interpret this as evidence that our simulation is more reactive, and suppresses unrealistic planner behavior more effectively—particularly in collision scenarios. Qualitative examples (see \cref{fig:policy_evaluation} and supplementary videos) further highlight that \ourmethod agents respond naturally in diverse scenarios, avoiding collisions when reasonable and minimizing unrealistic off-road behaviors.

For throughput, we benchmark against SMART \cite{wu2024smart}, a leading multi-agent motion generator on the WOSAC leaderboard \cite{montali2023wosac}. To ensure fairness, we run both methods at 5 Hz for full 8-second episodes on a single NVIDIA A100 GPU. SMART achieves $22.5 \pm 0.01$ scenarios/sec, while our method reaches $211.8 \pm 5.64$ scenarios/sec—a $\sim 10\times$ speedup. Moreover, GPUDrive can be optimized for another order-of-magnitude efficiency gain~\cite{cusumano2025robust, pufferai}. All experiments were run on a dual Intel Xeon Platinum 8358 (64 cores / 128 threads, 2.6 GHz) server with a single A100 GPU, and results are averaged over 5 seeds.

In summary, \ourmethod agents deliver both \textit{more realistic closed-loop evaluation} and \textit{significantly higher throughput}, enabling reliable and scalable benchmarking of ego-motion planning policies.

\section{Discussion}

\textbf{Limitation of WOSAC Metrics.} During experiments, we found that WOSAC metrics can produce misleading scores. In \cref{fig:wosac_fail}, the logged agent turns into a parking lot, while \ourmethod continues straight without crossing the curb. Although this behavior has an off-road rate of $0.0$, WOSAC penalizes it with a low map score because the metric rewards reproducing the logged trajectory rather than recognizing alternative valid behaviors. Likewise, when logs contain sensor noise leading to collisions or off-road trajectories, WOSAC assigns higher likelihood to agents that repeat these errors~\cite{wang2025llm}. We provide more qualitative results in \cref{supp:wosac_fail}. This reveals a key limitation: WOSAC evaluates similarity to logged distributions, not necessarily safety or human-likeness, and can misalign with the goals of self-play RL.
% distribution plots, show a few failure cases where agents cross the road,
% Also include a discusssion why RL and WOSAC metrics are not perfectly alinged. A few failure case on WOSAC metrics , show it in figure?

% As also mentioned in \cite{wang2025llm
\textbf{Simulating VRUs (Pedestrians and Cyclists).} 
SPACeR improves pedestrian and cyclist realism compared to prior baseline in a large margin, showing that self-play anchoring transfers beyond vehicle agents \cref{sec:exp:humanlike}. Nevertheless, VRU self-play remains underexplored: current WOSAC metrics (e.g., collision, off-road) are tailored to vehicles and do not capture sidewalk adherence, crosswalk usage, or other pedestrian specific behaviors. Reward design must likewise reflect these VRU-specific considerations. In this work we primarily use likelihood and collision signals for VRU motion. Developing VRU-aware metrics, reward shaping, and scene-level infrastructure is an important direction for enabling more realistic pedestrian and cyclist simulation.

\begin{wrapfigure}{r}{0.45\textwidth}
  \vspace{-10pt} % tighten space above the figure
  \centering
  \includegraphics[width=\linewidth]{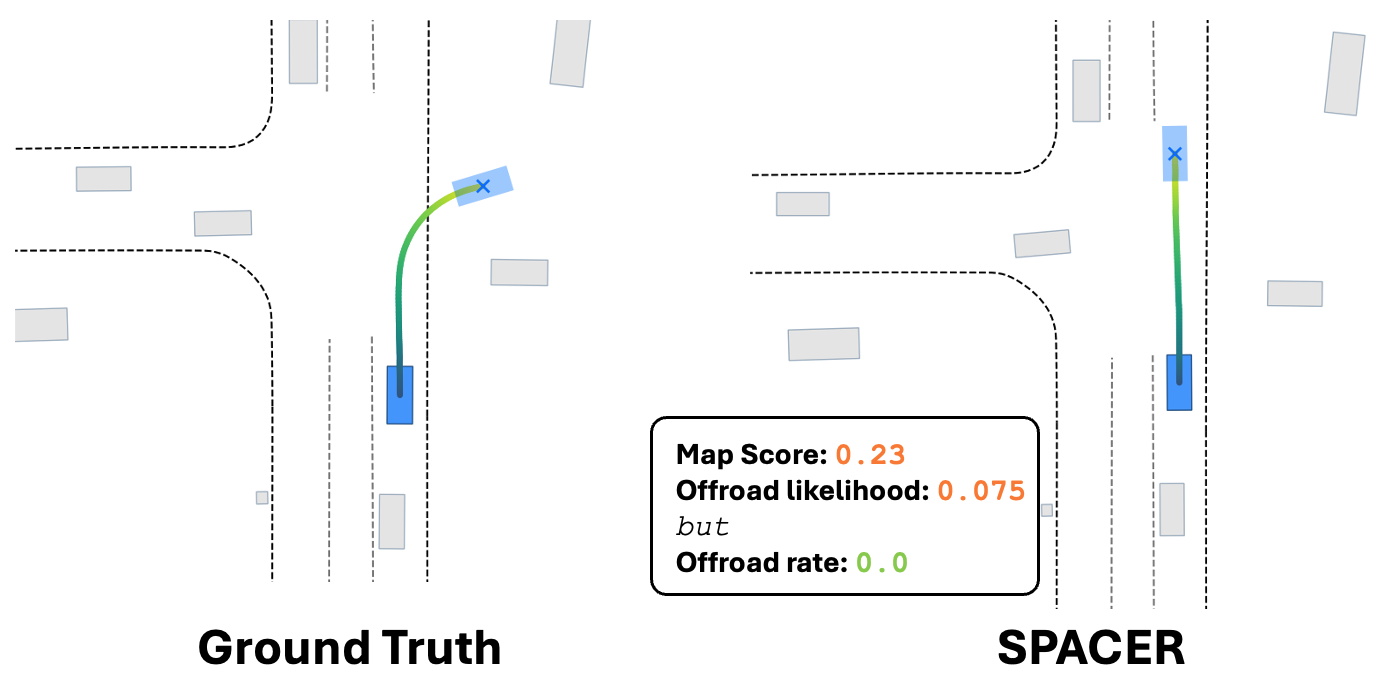}
  \vspace{-8pt} % tighten space between figure and caption
  \caption{\textbf{Example scenarios where WOSAC metrics produce unrealistic estimates}}
  \label{fig:wosac_fail}
  \vspace{-20pt} % tighten space after the figure
\end{wrapfigure}

\textbf{Training Efficiency Bottlenecks.} 
In our current setting, each run requires roughly 24–48 hours, partly due to GPUDrive’s lack of multi-GPU support. 
Future extensions could exploit multi-GPU training or alternative backends such as PufferLib~\cite{pufferai}, 
which has demonstrated order-of-magnitude speedups. Memory usage also limits scalability, especially for 
architectures like SMART that encode pairwise interactions explicitly. Recent advances in memory-efficient 
design (e.g.,~\cite{zhao2025drope}) may help reduce overhead and increase training throughput.

\section{Conclusion}
We introduced \ourmethod, which anchors self-play RL to a pretrained tokenized model via KL alignment, 
providing lightweight realism signals without relying on logged trajectories. On the Waymo Sim Agent Challenge, 
it achieves higher realism than prior self-play while producing policies that are $\sim$50$\times$ smaller 
and $\sim$10$\times$ faster than state-of-the-art imitation models. Our policies remain reactive and humanlike, avoiding the unrealistic collisions often observed in imitation methods during closed-loop planner evaluation. Together, these results position \ourmethod as a step toward scalable, real-time closed-loop evaluation, 
and ultimately training, of planners under realistic large-scale traffic scenarios.

\textbf{Reproducibility Statement.} We provide detailed implementation and training settings in Appendix~A.1--A.2.  Our approach are included in Sections~3.2, 3.3, and 4.1 to ensure clarity and reproducibility.

\section*{Acknowledgment}

This work was part of W.J. Chang’s summer internship at
Applied Intuition, and he is also supported by the National Science Foundation Graduate Research Fellowship
Program under Grant No. DGE 2146752. Any opinions,
findings, and conclusions or recommendations expressed
in this material are those of the author(s) and do not necessarily reflect the views of the National Science Foundation. The authors would like to thank Eric and Andrew for infrastructure support.

\maketitle

\bibliography{iclr2026_conference}
\bibliographystyle{iclr2026_conference}

\appendix
\clearpage

\section{Appendix}
\renewcommand{\thefigure}{A\arabic{figure}}
\renewcommand{\thetable}{A\arabic{table}}
\setcounter{figure}{0}
\setcounter{table}{0}
% \subsection{Reference Model Quality effect on SPACeR}

\subsection{Details on SPACeR design choices for simulating VRU}\label{supp:vru_details}
{
To extend SPACeR to simulating VRUs (pedestrians and cyclists), we ablate several design choices, as shown in \cref{tab:ablation_vru_metrics}.
We use separate action heads for each agent type and include a one-hot agent-type indicator in the ego observation so the policy can specialize its behavior.
We also find that the reference KL loss consistently improves VRU realism, helping reduce unstable or abrupt pedestrian behaviors.
Finally, both the goal-reaching loss and the multi-action head contribute to stable and human-like VRU motion, with noticeable drops in kinematic and interaction realism when either component is removed.}

{
For evaluation, all agents in the scene (including vehicles) are simulated, but WOSAC metrics are computed exclusively for pedestrian and cyclist targets. For training, we follow the same approach as \cite{wu2024smart}, where different agent types use their own token vocabularies.
}
\begin{table}[t]
\centering
\caption{{Ablation study on key components of the SPACeR adapted to VRU simulation settings. 
Each row removes one design choice. Metrics are reported over VRU target agents only.}}

\begin{tabular}{lccccc}
\toprule
\textbf{Ablation} 
& \textbf{Composite $\uparrow$} 
& \textbf{Kinematic $\uparrow$} 
& \textbf{Interactive $\uparrow$} 
& \textbf{Map-based $\uparrow$} 
& \textbf{minADE $\downarrow$} \\
\midrule
Full Model 
& \textbf{0.729} & \textbf{0.413} & 0.762 & \textbf{0.866} & \textbf{2.066} \\

-- goal-reaching weight 
& 0.728 & 0.405 & \textbf{0.769} & 0.859 & 2.295 \\

-- multi-action head 
& 0.685 & 0.323 & 0.742 & 0.818 & 3.416 \\

-- reference KL loss 
& 0.607 & 0.222 & 0.626 & 0.804 & 12.844 \\
\bottomrule
\end{tabular}
\label{tab:ablation_vru_metrics}
\end{table}

\subsection{Anchoring parameters ablation study}
In the ablation study \cref{tab:wosac_ablate}, KL alignment contributes most to realism. While log-likelihood rewards are
popular Escontrela et al. (2023), their impact on realism is modest—likely because the real-world driving
distribution is highly multi-modal, so maximizing likelihood alone does not yield a stable or sufficiently
discriminative signal. A second insight is that, once the policy is anchored to the reference distribution, the
explicit goal reward can be removed, which further improves realism.

{In \cref{fig:weight_ablation}, we visualize the training dynamics of the anchoring terms. Likelihood-only optimization steadily increases the log-likelihood but also drives entropy downward, indicating reduced diversity as training progresses. In contrast, adding KL alignment not only increases log-likelihood but also keeps the KL term stable and preserves higher entropy, preventing collapse and supporting more reliable top-k realism.}
% Needs: \usepackage{booktabs}
\begin{table}[t]
\centering
\footnotesize
\setlength{\tabcolsep}{2pt}
\renewcommand{\arraystretch}{1.05}
\begin{tabular}{@{}lcc@{}}
\toprule
\textbf{Variant} & \textbf{Composite.$\uparrow$} & \textbf{minADE$\downarrow$} \\
\midrule
$r_{\text{task}}$ only & 0.70 & 14.43 \\
Goal + LLH & 0.69 & 21.05 \\
Goal + KL & 0.73 & \textbf{4.08} \\
\midrule
KL + $r_{\text{inf}}$ & \textbf{0.74} & 4.73 \\
KL + $r_{\text{inf}}$ + LLH & \textbf{0.74} & 4.68 \\
\bottomrule
\end{tabular}
\caption{\textbf{Ablations on WOSAC (validation).} 
$r_{\text{inf}}$ = infraction penalties (off-road, collision); 
LLH = log-likelihood reward; 
KL alignment essential for realism; goals unnecessary.}
\label{tab:wosac_ablate}
\end{table}

\begin{figure}[t]
  \centering

  % ---------------- Top row ----------------
  \includegraphics[width=0.32\linewidth]{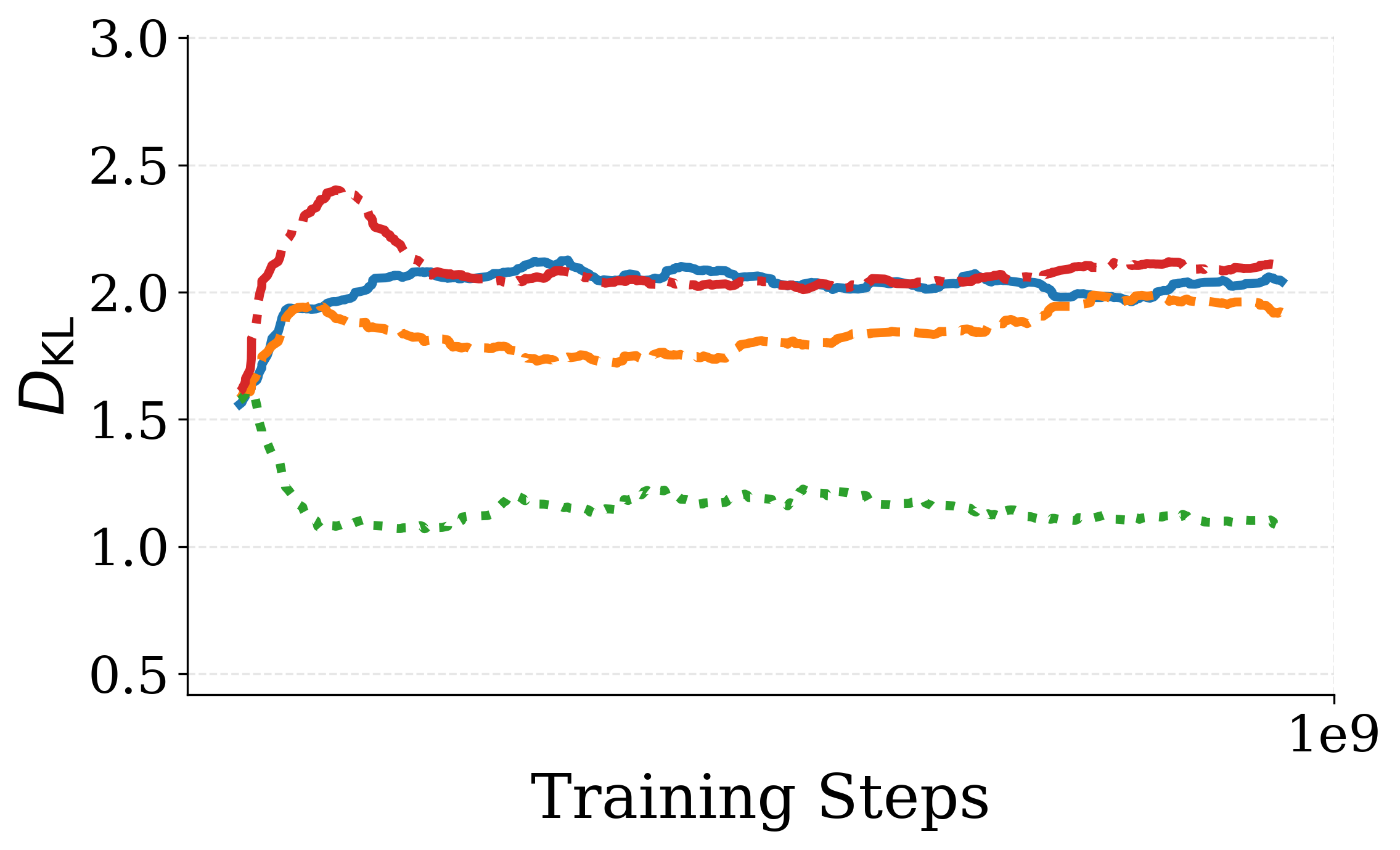}
  \hspace{0.3em}
  \includegraphics[width=0.32\linewidth]{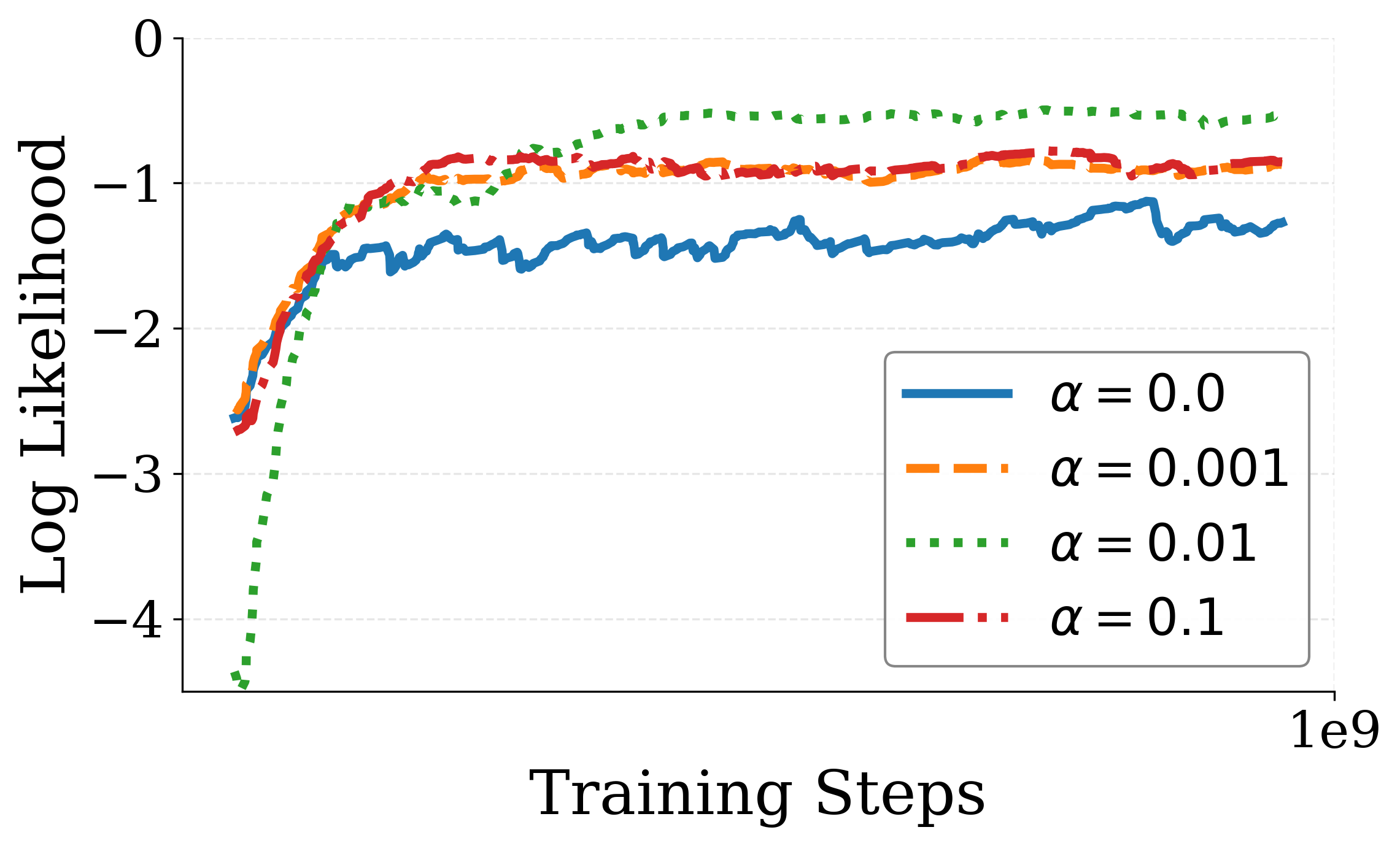}
  \hspace{0.3em}
  \includegraphics[width=0.32\linewidth]{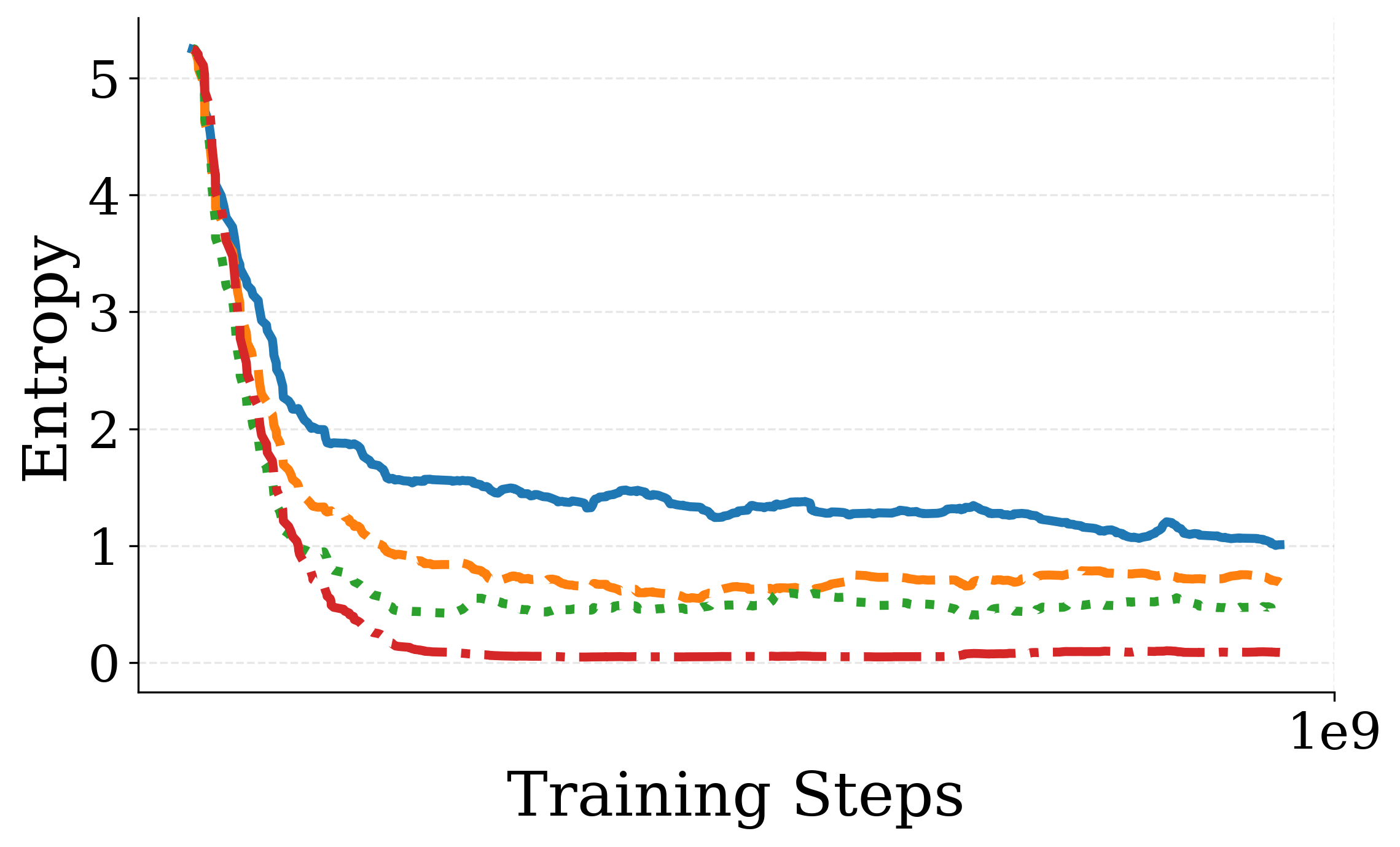}
  \makebox[\linewidth]{\textbf{(a)}}

  \vspace{0.7em}

  % ---------------- Middle row ----------------
  \includegraphics[width=0.32\linewidth]{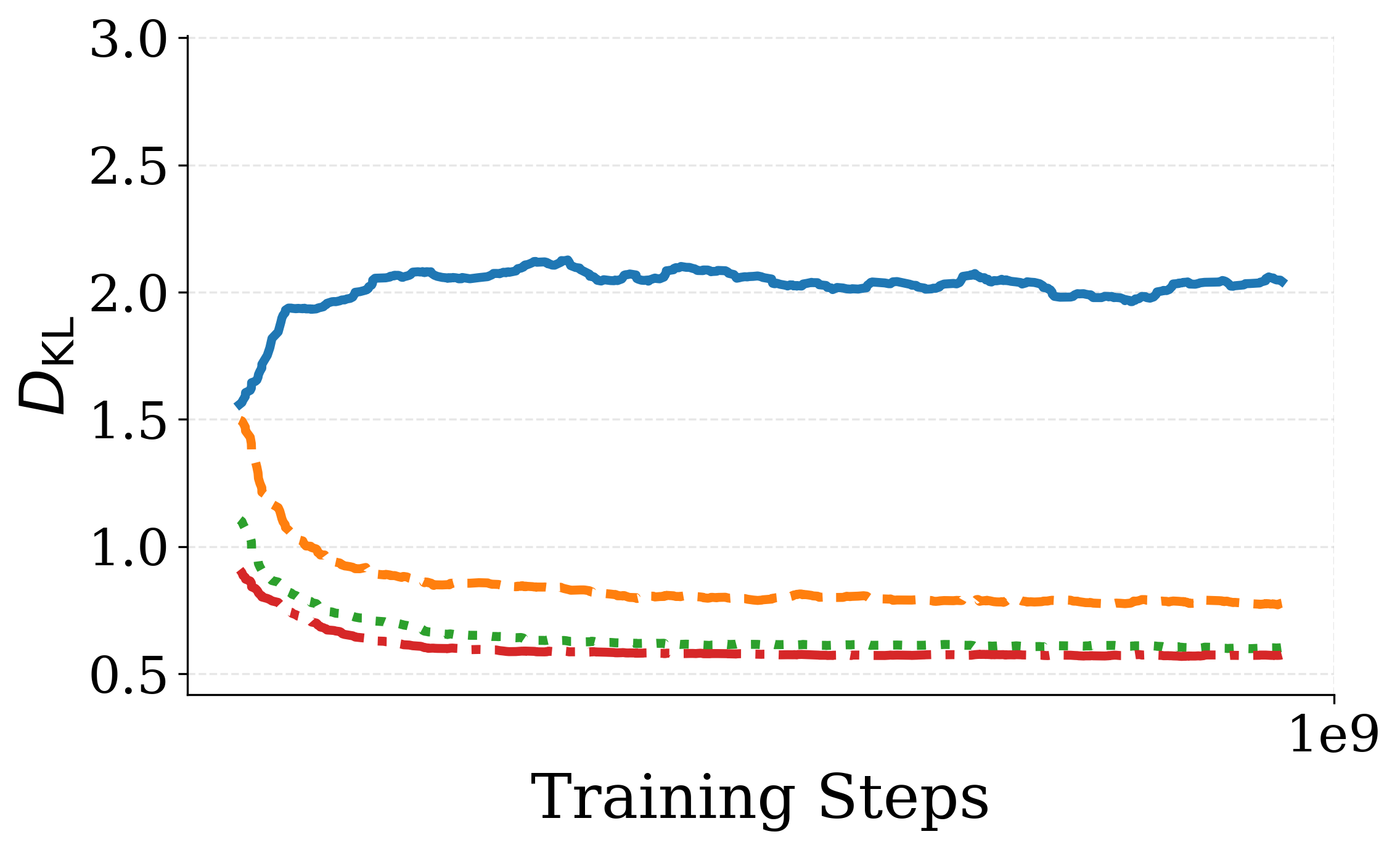}
  \hspace{0.3em}
  \includegraphics[width=0.32\linewidth]{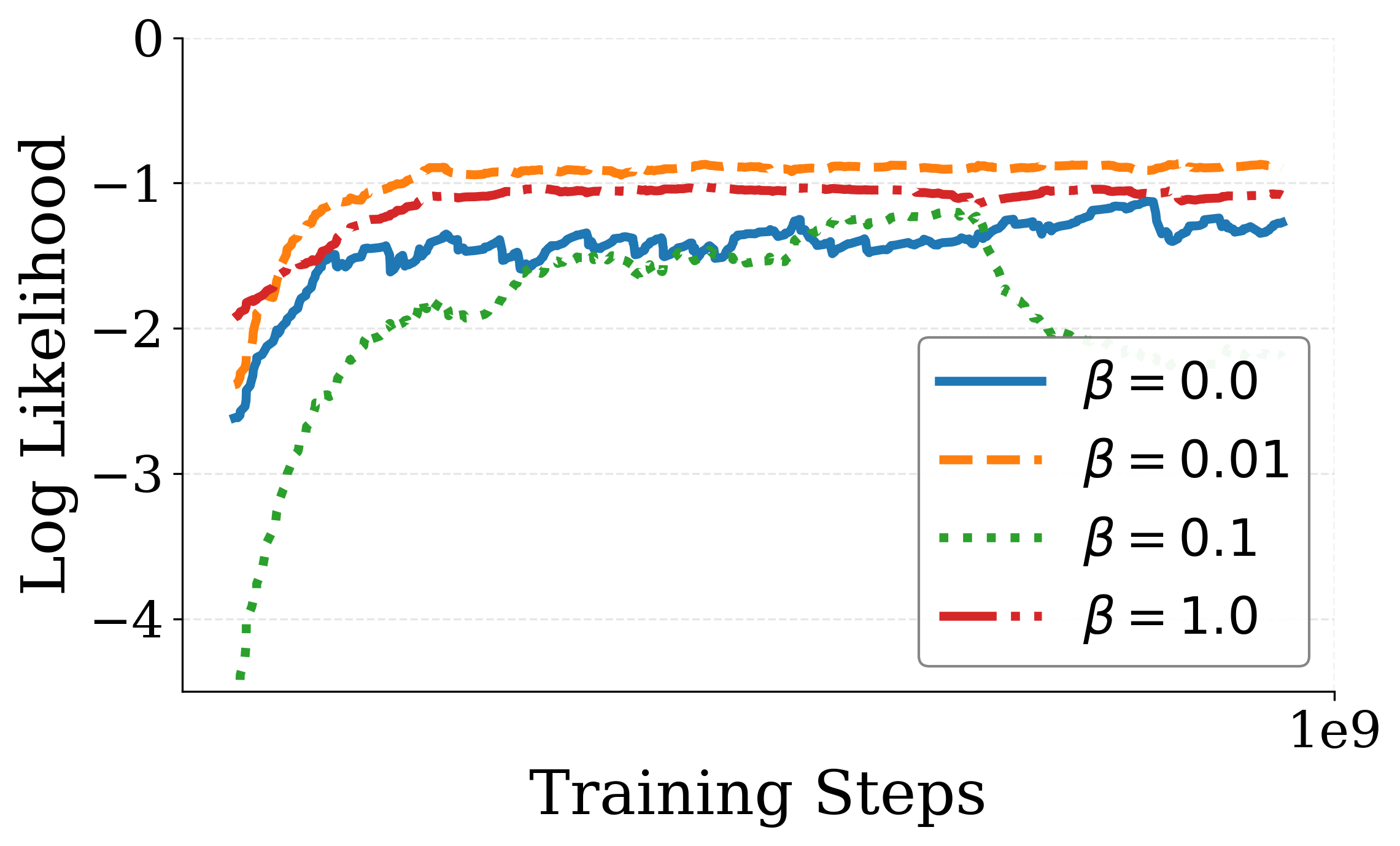}
  \hspace{0.3em}
  \includegraphics[width=0.32\linewidth]{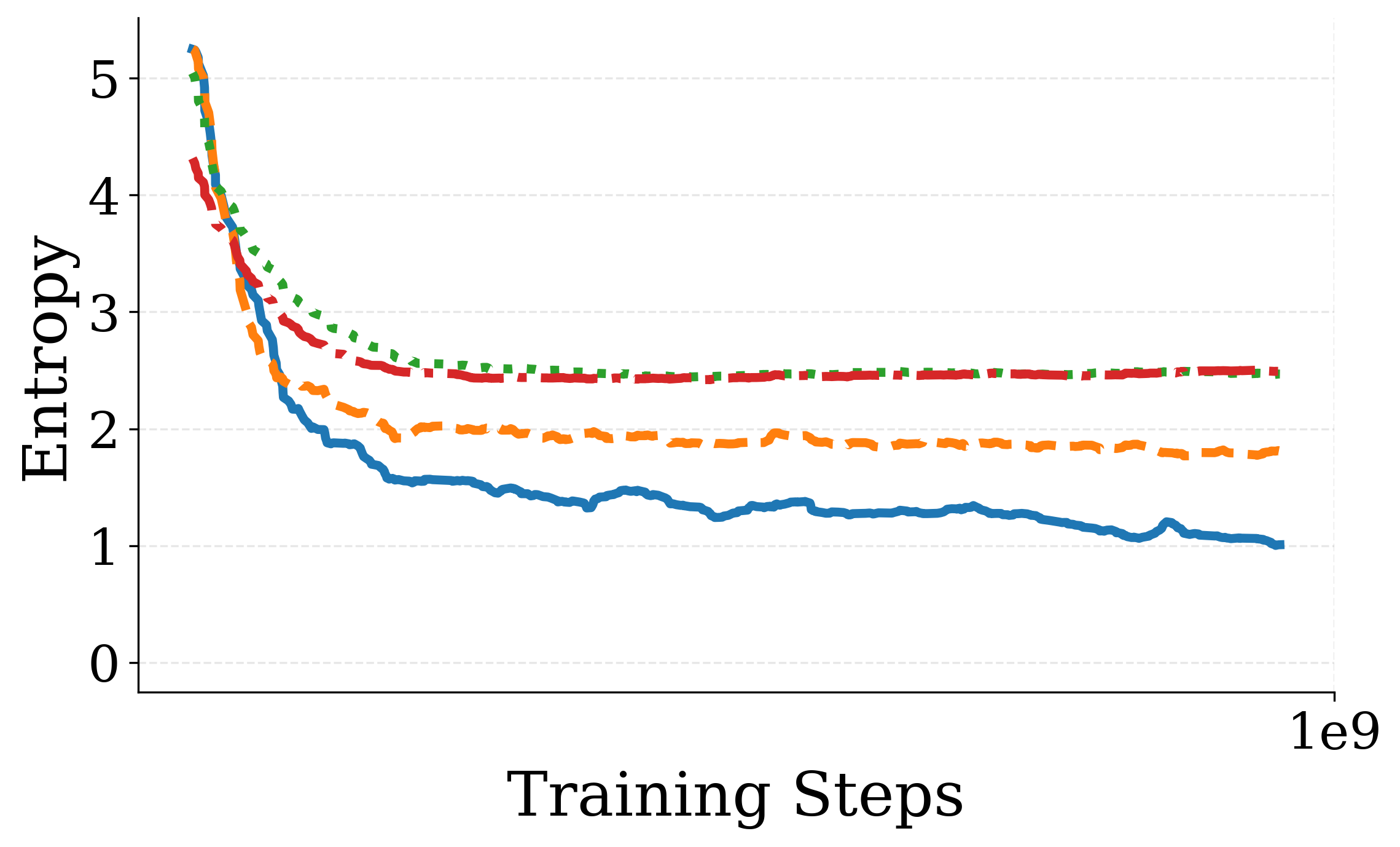}
  \makebox[\linewidth]{\textbf{(b)}}

  \vspace{0.7em}

  \caption{
    {
    \textbf{Anchoring parameters: effects of likelihood and KL-divergence weights.}
    (a) Likelihood weight effect without KL loss.
    (b) KL divergence weight sweep without likelihood loss.
    Likelihood-only optimization increases the probability of the most likely action but collapses entropy, while KL alignment maintains diversity and improves likelihood.}
}
  \label{fig:weight_ablation}
\end{figure}

% \textcolor{blue}{
% Additional EXP
% }

% \textbf{LLM usage section }
% LLMs were used to assist with writing, code implementation, and code review; all experimental design and final decisions were made by the authors.

% \textbf{Qualitative Videos.}
% We recommend readers to view the qualitative videos provided in the supplementary materials.

\subsection{Training hyperparamters and implementation details of Self-play}\label{sup:training_details}
The core PPO hyperparameters are summarized in   \cref{tab:ppo-hparams}. 
When training without a reference model, we employ 600 parallel worlds on a single A100 (80\,GB, PCIe) GPU. 
For human-regularized PPO and reference-model training, we reduce the number of worlds to 300, which leads to an approximately $2\times$ slowdown due to the lack of multi-GPU support in \textsc{GPUDRIVE} with madrona backend. 
With multi-GPU support, we would expect comparable throughput to the reference-free setting. 
Our policy/value network is configured with an input embedding dimension of 64, a hidden dimension of 128, and a dropout rate of 0.01. 
To reduce GPU memory consumption in the reference-model setting, we highlight two adjustments: 
(1) we cap the maximum number of unique scenarios per batch at 200 to increase training speed, and 
(2) we limit the maximum number of map elements per agent from 200~\cite{kazemkhani2024gpudrive} to 120.
\cref{tab:ppo-hparams}

\begin{table}[t]
\centering
\caption{PPO Training Hyperparameters.}
\label{tab:ppo-hparams}
\begin{tabular}{lll}
\toprule
\textbf{Parameter} & \textbf{Value} & \textbf{Description} \\
\midrule
seed                & 42            & Random seed. \\
total\_timesteps    & 1{,}000{,}000{,}000 & Total number of environment timesteps. \\
batch\_size         & 131{,}072     & Timesteps collected per rollout. \\
minibatch\_size     & 8{,}192       & Timesteps per optimization minibatch. \\
learning\_rate      & 3e-4          & Optimizer learning rate. \\
anneal\_lr          & false         & Learning rate annealing. \\
gamma               & 0.99          & Discount factor. \\
gae\_lambda        & 0.95          & GAE parameter $\lambda$. \\
update\_epochs      & 4             & Optimization epochs per rollout. \\
norm\_adv           & true          & Normalize advantages. \\
clip\_coef          & 0.2           & PPO policy clip coefficient. \\
clip\_vloss         & false         & Clip value loss. \\
vf\_clip\_coef      & 0.2           & Value function clipping coefficient. \\
ent\_coef           & 0.0001        & Entropy coefficient. \\
vf\_coef            & 0.3           & Value loss coefficient. \\
max\_grad\_norm     & 0.5           & Gradient clipping (max L2 norm). \\
\bottomrule
\end{tabular}
\end{table}

\paragraph{Human-Regularized PPO Baseline.} 
We follow the setup of human-regularized PPO~\cite{cornelisse2024human}, but adapt it to our tokenized trajectory action space instead of low-level control actions. 
To train the behavior cloning (BC) reference policy, we use observation–action pairs from the full Waymo Open Motion Dataset (WOMD) rather than the 200-scenario subset originally used, ensuring a stronger expert baseline. 
Our BC model is parameterized with roughly $2\times$ the capacity of the self-play policy network to increase expressiveness, achieving a validation accuracy of 92\%. 
We implement BC training using the \texttt{imitation} package~\cite{gleave2022imitation} and train for 60 epochs on the full dataset.

During HR-PPO training, we regularize the learned policy against the BC reference policy using a KL-divergence penalty with weight $\beta=0.01$. 
Larger values of $\beta$ destabilize training, while using tokenized model can generally increase to $\beta=1.0$ w/ similar performance. 
In addition, when applying human-regularized PPO in the original action space, we found it necessary to clamp the minimum reference log-probability to $10^{-20}$ to avoid unstable gradients; interestingly, this instability does not occur when using our tokenized trajectory reference model.

In the first submission, HR-PPO used the reverse KL divergence, the same as SPACeR. 
In the camera-ready version, we use the forward KL divergence and report the better-performing variant, without affecting the main conclusions.

\subsection{Training Details of SMART and CatK.}
We pretrain the base \textsc{SMART} model on 16$\times$A100 (80\,GB, PCIe) GPUs for 10 epochs and select the checkpoint with the best validation loss ($5\times 10^{-4}$). 
We then apply closed-loop supervised finetuning as described in~\cite{zhang2024closed_catk} for 6 additional epochs with an effective batch size of 64 and a learning rate of $1\times 10^{-5}$. 
The final model contains 3.2M parameters and is trained on the full training dataset of approximately 500k scenarios. 
To make the self-play policy more responsive while maintaining a balance between control frequency and memory usage, we use a sampling frequency of 5\,Hz instead of the original 2\,Hz.

% \subsection{Throughput Benchmarking Details}

% To ensure a fair comparison, we run both SMART \cite{wu2024smart} and our approach at 5 Hz for full 
% 8-second episodes on a single NVIDIA A100 GPU. 
% SMART achieves $22.5 \pm 0.01$ scenarios/sec, while our method reaches $211.8 \pm 5.64$ scenarios/sec—a 
% $\sim 10\times$ speedup. 
% Moreover, GPUDrive can be further optimized for another order-of-magnitude efficiency gain 
% \cite{cusumano2025robust, pufferai}. 
% All experiments were conducted on a dual Intel Xeon Platinum 8358 
% (64 cores / 128 threads, 2.6 GHz) server with a single A100 GPU, 
% and results are averaged over 5 seeds.

\subsection{Waymo Sim Agent Challenge Metrics}\label{sup:metrics}

To evaluate whether the trained policies are humanlike, we follow the evaluation protocol of the Waymo Open Sim Agent Challenge (WOSAC)\cite{montali2023wosac}, which measures how closely the distribution of simulated rollouts matches the ground-truth distribution across kinematics, agent interactions, and map adherence.

The WOSAC metrics quantify how closely the distribution of simulated rollouts matches the ground-truth distribution, across multiple aspects such as kinematics, agent interactions, and map adherence. 

Concretely, for each scenario containing up to 128 agents simulated for 8 seconds, we generate 32 multi-agent rollout samples. For a target agent $a$ in scenario $i$ and statistic $F_j$, the negative log-likelihood (NLL) of the ground-truth outcome under the empirical distribution of simulated samples is defined as:
\begin{equation}
\text{NLL}(i,a,t,j) = -\log \, p_{i,j,a}\!\left(F_j(x^*(i,a,t))\right),
\end{equation}
where $p_{i,j,a}(\cdot)$ denotes the empirical distribution constructed from the simulated samples, and $x^*(i,a,t)$ is the true trajectory at time $t$. Lower values indicate that the simulation better reflects observed behavior.

To obtain a per-agent summary, we aggregate over valid timesteps:
\begin{equation}
m(a,i,j) = \exp \Bigg(- \frac{1}{N(i,a)} \sum_{t} v(i,a,t)\,\text{NLL}(i,a,t,j) \Bigg),
\end{equation}
where $N(i,a) = \sum_{t} v(i,a,t)$ is the number of valid timesteps for agent $a$. The scenario-level score is then computed as the average across all evaluated agents:
\begin{equation}
m(i,j) = \frac{1}{A_{\text{target}}} \sum_{a} m(a,i,j),
\end{equation}
with $A_{\text{target}}$ denoting the number of target agents in the scenario.

In our experiments, we restrict evaluation to vehicles only: agents corresponding to pedestrians or cyclists are excluded as targets and fixed to their ground-truth trajectories. All reported results are computed on a 2\% validation subset of the dataset \cite{zhang2024closed_catk}.

% \begin{figure}
%     \centering
%     \includegraphics[width=1\linewidth]{fig/likelihood.png}
%     \caption{Enter Caption}
%     \label{fig:placeholder}
% \end{figure}
\subsection{WOSAC Failure Qualitative Results}\label{supp:wosac_fail}
More qualitative results of WOSAC limitation provided in \cref{fig:wosac_fail_supp}
\begin{figure}[t]
    \centering
    \includegraphics[width=0.7\linewidth]{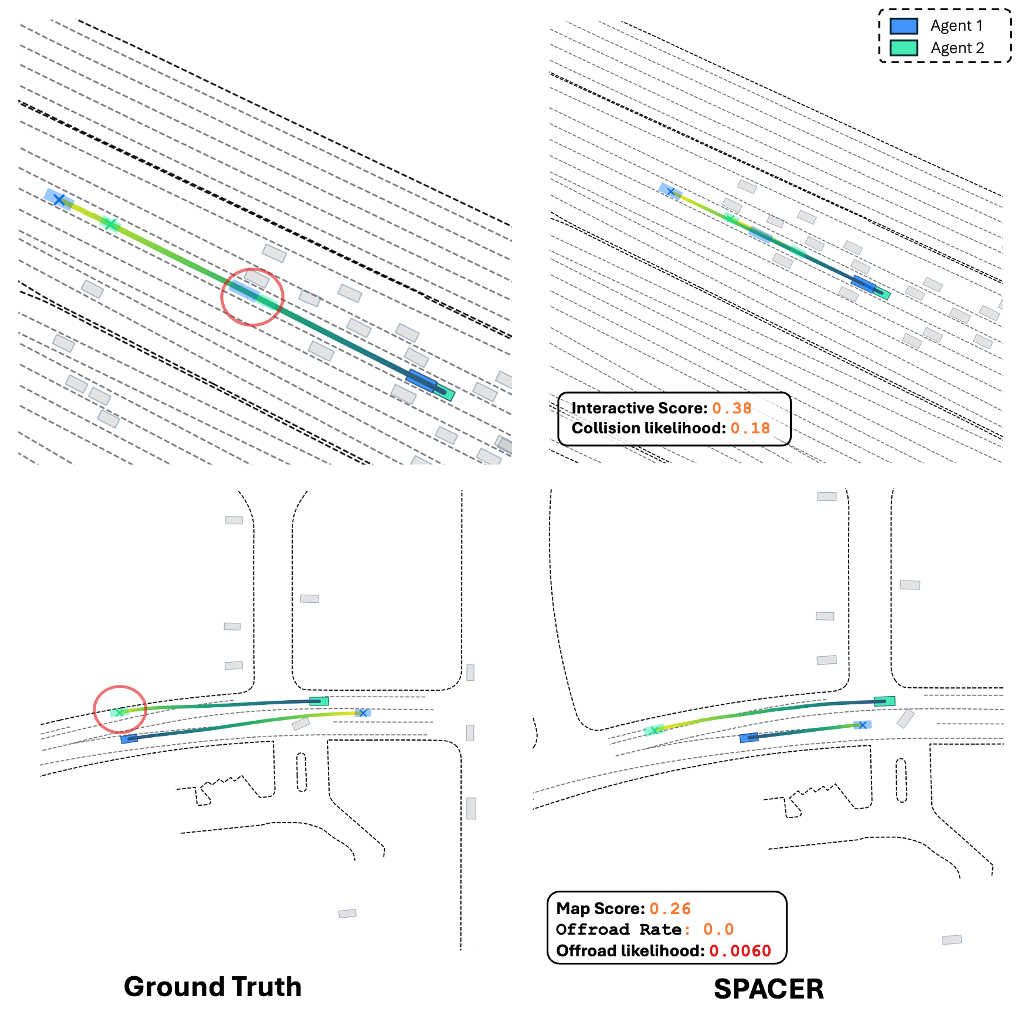}
      \caption{\textbf{Additional qualitative examples of WOSAC limitations.} 
    Top row: due to sensor noise, the ground-truth agents collided in the data; 
    consequently, safe simulated behaviors are assigned low collision likelihood.  
    Bottom row: the ground-truth agent collided with the road edge, so remaining 
    on-road is given near-zero off-road likelihood and a low map score. 
    These cases illustrate how WOSAC can penalize safe and realistic behaviors 
    when they diverge from noisy or imperfect logged trajectories.}
    \label{fig:wosac_fail_supp}
\end{figure}

\subsection{Planner Variants Overview}\label{sup:planner_variant}

Our evaluation framework covers two categories: \textbf{learning-based} and \textbf{rule-based} planners.  

\textbf{Learning-based.} We trained 22 self-play reinforcement learning policies in GPUDrive 
\cite{kazemkhani2024gpudrive}, all sharing the same decentralized late-fusion MLP architecture 
(controlling up to 64 agents) but differing in reward weights (see Sec.~A.5.1).  

\textbf{Rule-based.} We include two families of classical planners: Frenet-based trajectory 
samplers and Intelligent Driver Model (IDM) planners, each with 10 parameterized variants 
ranging from conservative to aggressive driving styles.

\subsubsection{Self-Play Policies}

In addition to handcrafted planners, we trained 22 self-play policies using 
the GPUDrive framework \cite{kazemkhani2024gpudrive}. 
All policies share the same decentralized late-fusion MLP architecture 
(controlling up to 64 agents), but differ in reward weighting.  

We varied the weights of three components: goal-reaching 
($w_{\text{goal}} \in \{1.0, 0.5\}$), collision penalties 
($w_{\text{collided}} \in \{0, -0.375, -0.75, +0.1\}$), and off-road penalties 
($w_{\text{offroad}} \in \{0, -0.375, -0.75, +0.1\}$). 
This grid covers both standard reward shaping (non-negative penalties) 
and unconventional settings with negative weights, where agents are 
encouraged to collide or go off-road.  

The resulting 22 policies span a wide spectrum of behaviors, from highly 
conservative (strong safety penalties) to adversarial (negative penalties), 
allowing us to stress-test planner evaluation under diverse multi-agent 
conditions. All policies are trained for 1B environment steps on 10k 
resampled WOMD scenarios, with 600 parallel worlds and up to 64 controlled 
agents per rollout.

\subsubsection{Frenet-based Planners}

The Frenet planner uses quintic/quartic polynomials to generate trajectory candidates in the Frenet frame, evaluating them based on a weighted cost function that considers:
\begin{itemize}
    \item Lateral deviation from centerline ($w_{\text{lateral}}$)
    \item Velocity tracking ($w_{\text{velocity}}$)
    \item Acceleration smoothness ($w_{\text{acceleration}}$)
    \item Progress along path ($w_{\text{progress}}$)
    \item Jerk minimization ($w_{\text{jerk}}$)
    \item Collision avoidance (collision penalty)
\end{itemize}

Table~\ref{tab:frenet_variants} summarizes the 10 Frenet-based planner variants, showing their key characteristics including speed ranges, lateral weights, safety focus levels, and sampling densities. These variants range from conservative safety-focused configurations to aggressive high-speed optimized settings.

\begin{table}[h!]
\centering
\begin{threeparttable}
\caption{Summary of Frenet-based Planner Variants}
\label{tab:frenet_variants}
% \resizebox{\textwidth}{!}{%
\begin{tabular}{lllcccc}
\toprule
\textbf{Variant} & \textbf{Description} & \textbf{Speed} & \textbf{Lateral} & \textbf{Safety} & \textbf{Sampling}\tnote{a} & \textbf{Key Features} \\
 & & \textbf{(m/s)} & \textbf{Weight} & \textbf{Focus} & \textbf{(d,v,t)} & \\
\midrule
Baseline & Balanced & 0-30 & 10.0 & Medium & 10,5,3 & Standard configuration \\
Aggressive & High progress & 0-35 & 5.0 & Low & 10,5,3 & Progress weight = 2.0 \\
Conservative & Safety-first & 0-20 & 50.0 & High & 10,5,3 & Collision penalty = 5000 \\
Smooth Rider & Comfort & 0-30 & 20.0 & Medium & 10,5,3 & Jerk weight = 3.0 \\
Lane Keeper & Centerline & 0-30 & 100.0 & Medium & 15,5,3 & Lateral span = 1.5m \\
Wide Search & Comprehensive & 0-30 & 10.0 & Medium & 20,10,7 & Large search space \\
Fast Planner & Quick & 0-30 & 10.0 & Medium & 5,3,2 & Reduced horizon \\
Long Horizon & Strategic & 0-30 & 10.0 & Medium & 10,5,3 & 40 horizon steps \\
No Collision & Test baseline & 0-30 & 10.0 & None & 10,5,3 & Collision disabled \\
High Speed & Highway & 5-40 & 10.0 & Medium & 10,5,3 & Velocity span = 15 \\
\bottomrule
\end{tabular}%
% } %
\begin{tablenotes}
\small
\item[a] Sampling notation: (d,v,t) represents (lateral samples, velocity samples, time samples)
\end{tablenotes}
\end{threeparttable}
\end{table}

\subsubsection{IDM-based Planners}

The IDM planner implements the Intelligent Driver Model for longitudinal control combined with a PID controller for lateral tracking. Key parameters include:
\begin{itemize}
    \item Desired velocity ($v_0$)
    \item Minimum spacing ($s_0$)
    \item Safe time headway ($T$)
    \item Maximum acceleration ($a$)
    \item Comfortable deceleration ($b$)
    \item Aggressiveness factor (0.0-1.0)
\end{itemize}

Table~\ref{tab:idm_variants} presents the 10 IDM-based planner variants, each configured to represent different driving styles from cautious urban driving to aggressive highway scenarios. The aggressiveness factor plays a key role in determining the overall behavior of each variant.

\begin{table}[h!]
\centering
\begin{threeparttable}
\caption{Summary of IDM-based Planner Variants}
\label{tab:idm_variants}
% \resizebox{\textwidth}{!}{%
\begin{tabular}{llccccc}
\toprule
\textbf{Variant} & \textbf{Description} & \textbf{Desired} & \textbf{Min Gap} & \textbf{Headway} & \textbf{Aggress.}\tnote{b} & \textbf{Special} \\
 & & \textbf{Vel (m/s)} & \textbf{$s_0$ (m)} & \textbf{$T$ (s)} & \textbf{Factor} & \textbf{Features} \\
\midrule
IDM Baseline & Standard & 30 & 2.0 & 1.5 & 0.5 & Balanced behavior \\
IDM Conservative & Cautious & 25 & 3.0 & 2.0 & 0.2 & Safety factor = 1.5 \\
IDM Aggressive & Dynamic & 35 & 1.5 & 1.0 & 0.8 & Safety factor = 0.9 \\
IDM Comfort & Smooth & 28 & 2.5 & 1.8 & 0.3 & Max jerk = 2.0 \\
IDM Highway & High-speed & 40 & 3.0 & 1.2 & 0.6 & Perception = 100m \\
IDM City & Urban & 15 & 2.0 & 1.5 & 0.4 & Perception = 30m \\
IDM Truck & Heavy & 25 & 4.0 & 2.0 & 0.3 & Length = 8.0m \\
IDM Emergency & Urgent & 40 & 1.5 & 0.8 & 0.9 & Max accel = 4.0 \\
IDM Adaptive & Balanced & 30 & 2.5 & 1.5 & 0.5 & Reaction = 0.2s \\
IDM Defensive & Safety & 25 & 4.0 & 2.5 & 0.1 & TTC\tnote{c} = 3.0s \\
\bottomrule
\end{tabular}%
% }
\begin{tablenotes}
\small
\item[b] Aggressiveness factor: Ranges from 0.0 (very conservative) to 1.0 (very aggressive)
\item[c] TTC: Time-to-collision threshold
\end{tablenotes}
\end{threeparttable}
\end{table}

Table~\ref{tab:config_comparison} provides a detailed comparison of the key configuration parameters for representative variants from both planner types, highlighting the differences in their weight distributions and fundamental parameters that lead to their distinct behaviors.

\begin{table}[h!]
\centering
\begin{threeparttable}
\caption{Key Configuration Parameters Comparison}
\label{tab:config_comparison}
\begin{tabular}{lccccc}
\toprule
\textbf{Parameter} & \textbf{Baseline} & \textbf{Aggressive} & \textbf{Conservative} & \textbf{Smooth} & \textbf{Lane Keeper} \\
\midrule
\multicolumn{6}{c}{\textit{Frenet Planner Weights}} \\
\midrule
Lateral ($w_l$) & 10.0 & 5.0 & 50.0 & 20.0 & 100.0 \\
Velocity ($w_v$) & 1.0 & 0.5 & 1.0 & 2.0 & 1.0 \\
Acceleration ($w_a$) & 1.0 & 1.0 & 3.0 & 5.0 & 1.0 \\
Progress ($w_p$) & 1.0 & 2.0 & 1.0 & 1.0 & 1.0 \\
Jerk ($w_j$) & 0.5 & 0.5 & 1.5 & 3.0 & 0.5 \\
\midrule
\multicolumn{6}{c}{\textit{IDM Parameters}} \\
\midrule
Desired vel ($v_0$) & 30.0 & 35.0 & 25.0 & 28.0 & - \\
Min spacing ($s_0$) & 2.0 & 1.5 & 3.0 & 2.5 & - \\
Time headway ($T$) & 1.5 & 1.0 & 2.0 & 1.8 & - \\
Max accel ($a$) & 2.0 & 3.0 & 1.5 & 1.5 & - \\
Comfort decel ($b$) & 3.0 & 4.0 & 2.0 & 2.0 & - \\
\bottomrule
\end{tabular}%
\begin{tablenotes}
\small
\item Note: All planners use dt=0.1s time step and wheelbase=2.8m
\end{tablenotes}
\end{threeparttable}
\end{table}

\end{document}